\documentclass[a4paper,twocolumn]{esapub2005} 
\usepackage{times}
\usepackage{natbib}
\usepackage{url}
\usepackage{graphicx, xcolor, caption, subcaption, tikz, pgfplots, colortbl, booktabs}
\usetikzlibrary{positioning, calc, patterns,decorations.pathreplacing,3d,math,pgfplots.fillbetween, matrix}

\definecolor{firstcolor}{RGB}{241, 213, 131}
\definecolor{secondcolor}{RGB}{216, 216, 216}
\definecolor{thirdcolor}{RGB}{245,194,71}

\title{On the geometric accuracy of implicit and primitive-based representations derived from view rendering constraints}
\author[1, 2]{Elias De Smijter}
\author[1, 3]{Renaud Detry}
\author[2]{Christophe De Vleeschouwer}
\affil[1]{KU Leuven, Dept. Electrical Engineering, Research unit Processing Speech and Images, B-3000 Leuven, Belgium}
\affil[2]{UCLouvain, ICTEAM, Dept. Electrical Engineering, B-1348 Ottignies-Louvain-la-Neuve, Belgium}
\affil[3]{KU Leuven, Dept. Mechanical Engineering, Research unit Robotics, Automation and Mechatronics, B-3000 Leuven, Belgium}

\begin{document}

\keywords{Novel View Synthesis; geometry estimation}

\maketitle

\begin{abstract}
We present the first systematic comparison of implicit and explicit Novel View Synthesis methods for space-based 3D object reconstruction, evaluating the role of appearance embeddings. While embeddings improve photometric fidelity by modeling lighting variation, we show they do not translate into meaningful gains in geometric accuracy — a critical requirement for space robotics applications. Using the SPEED+ dataset, we compare K-Planes, Gaussian Splatting, and Convex Splatting, and demonstrate that embeddings primarily reduce the number of primitives needed for explicit methods rather than enhancing geometric fidelity. Moreover, convex splatting achieves more compact and clutter-free representations than Gaussian splatting, offering advantages for safety-critical applications such as interaction and collision avoidance. Our findings clarify the limits of appearance embeddings for geometry-centric tasks and highlight trade-offs between reconstruction quality and representation efficiency in space scenarios.
\end{abstract}

%
%
%

\section{Introduction}
\label{sec:intro}
Novel View Synthesis (NVS) models such as neural radiance fields \cite{nerf} and 3D Gaussian Splatting \cite{gaussiansplatting} have demonstrated a remarkable ability to reconstruct both the appearance and geometry of a scene from a set of posed monocular images. Their success has spurred applications beyond photorealistic rendering, particularly in domains where geometric information is essential, such as Simultaneous Localization and Mapping (SLAM) \cite{loopsplat, gsslam}.

This geometric capability is, however, somewhat paradoxical: NVS models are optimized primarily with appearance-based loss functions, with no explicit enforcement of geometric constraints. As a result, their learned geometry often contains inconsistencies. In the case of Gaussian Splatting, for example, spurious structures — commonly referred to as \emph{floaters} — may appear in free space where no physical surface exists. While such artifacts are typically unobtrusive in rendering applications, they expose a fundamental tension between appearance-based training and geometric accuracy. This tension becomes critical in geometry-focused tasks such as SLAM and full 3D object reconstruction \cite{3dgstomesh,3dgsreview}.

The challenging conditions of geometry-focused tasks are especially pronounced in space-based object analysis, where targets appear against the uniform backdrop of space and often exhibit highly regular geometric structures. At the same time, the spatial domain presents a distinctive opportunity: unlike terrestrial scenarios, which typically allow only hemispherical views, space-based observations enable nearly full spherical coverage. This creates a novel but underexplored reconstruction regime, with both unique difficulties and potential advantages.

Within this setting, floater artifacts persist as a substantial limitation with multiple potential causes. This research specifically addresses how these artifacts manifest as a function of distinct model design parameters, examining the canonical scenario where no visual distractors are present in images to trigger additional floater generation. For space-based datasets, this methodology involves filtering out images where the Earth's disc is observable behind the spacecraft.

We focus on two model design factors in particular: an implicit or explicit architecture and the incorporation of appearance embeddings. To this end, we compare three architectures. K-Planes \cite{kplanes} encode scenes implicitly using hash tables and multilayer perceptrons. Gaussian Splatting \cite{gaussiansplatting} offers an explicit representation that has been widely adopted across NVS tasks. Convex Splatting \cite{convexsplatting} replaces Gaussians with convex primitives, making it especially well-suited for capturing the structured geometry of manufactured space objects. 

Previous work has shown that modeling lighting variations through appearance embeddings — originally introduced in NeRF in the Wild \cite{nerfinthewild} — improves rendering quality \cite{kplanes,nerfinthewild}, but their effect on geometric accuracy remains unclear. This gap motivates our study: methods optimized for photorealistic rendering may require significant adaptation before they can reliably serve geometry-critical applications such as space robotics. 

In summary, the contribution of this work is a systematic comparison of state-of-the-art NVS methods for geometry-centric reconstruction in space robotics scenarios. Specifically, we (i) show that the impact of appearance embeddings on the geometric fidelity is very limited, only reducing some floater artifacts, (ii) evaluate representative implicit and explicit architectures on the SPEED+ spacecraft dataset, and (iii) establish that adapting primitives to align with anticipated object geometries substantially decreases the number of primitives required. To our knowledge, this is the first study to examine the role of appearance embeddings in NVS-based reconstruction under the unique conditions of space imagery.

\section{Related Works}
\label{sec:relatedworks}
In previous works, some neural scene representations have been applied to space-based 3D object reconstruction \cite{nerfgrafspace, inerfdnerfspace, antoinespace}. These efforts have primarily relied on implicit methods derived from the original Neural Radiance Field (NeRF) formulation \cite{nerf}, which models scenes through fully connected neural networks. While successful in capturing appearance and geometry, implicit representations are computationally expensive and slow to train. In contrast, recent advances in explicit methods such as Gaussian Splatting \cite{gaussiansplatting} and Convex Splatting \cite{convexsplatting} offer superior rendering quality and efficiency, making them attractive candidates for real-time or resource-constrained applications such as space robotics.

The work in \cite{nerfgrafspace} investigated NeRF in its standard formulation and in an adversarial training variant that removes the need for pose supervision. While the adversarial model produced results close to supervised NeRF, it exhibited overfitting that degraded depth estimation quality. Both approaches also struggled with complex lighting phenomena such as shadows and reflections. Moreover, their reliance on synthetic datasets with fixed illumination fails to capture the variable solar lighting conditions typical in orbital scenarios.

Building on this, \cite{inerfdnerfspace} extended the comparison to include iNGP \cite{ingp}, a fast-training NeRF variant, and D-NeRF \cite{dnerf}, which models dynamic objects. This work leveraged hardware-in-the-loop imagery that better reflects real orbital conditions, though this methodological difference complicates direct comparison with \cite{nerfgrafspace}. Their results showed that iNGP achieved comparable rendering quality to NeRF with significantly faster training—a crucial advantage for spacecraft with limited onboard computation. By contrast, D-NeRF yielded inconsistent improvements, casting doubt on its practicality for space-based reconstruction. These observations motivate our choice for using K-Planes \cite{kplanes} - a fast implementation similar to iNGP - as the baseline for our study.

Both prior studies also rely on specialized datasets that have seen limited adoption, whereas our work employs the SPEED+ dataset \cite{speedplus}, a widely used benchmark in the space robotics community. In the following section, we outline our methodology for evaluating neural scene representations on this dataset. We begin by reviewing the preliminaries of the three models under consideration before detailing our experimental setup.


\section{Methodology}
\label{sec:methodology}

\begin{figure}
   \centering
   \resizebox{0.49\textwidth}{!}{\input{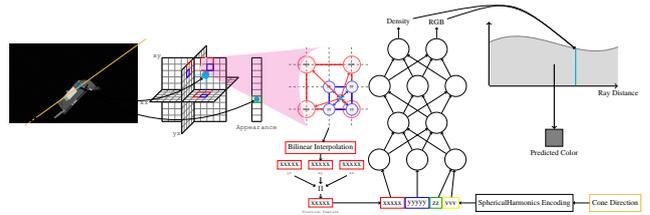}}
   \caption{K-Planes pipeline. In the first step, a feature vector is computed  by projecting the 3D  point position on three orthogonal planes supporting learnt feature maps, and by averaging the plane features in the projected positions across the multiple planes, considered at multiple resolutions. In the second step, the obtained feature vector is translated into color and density, uing an MLP. In the third step, these values get accumulated along a ray into a pixel value.}
   \label{fig:kplanes}
\end{figure}
\begin{figure}
   \centering
   \hfill
   \begin{subfigure}[b]{0.22\textwidth}
      \centering
      \includegraphics[width=\textwidth]{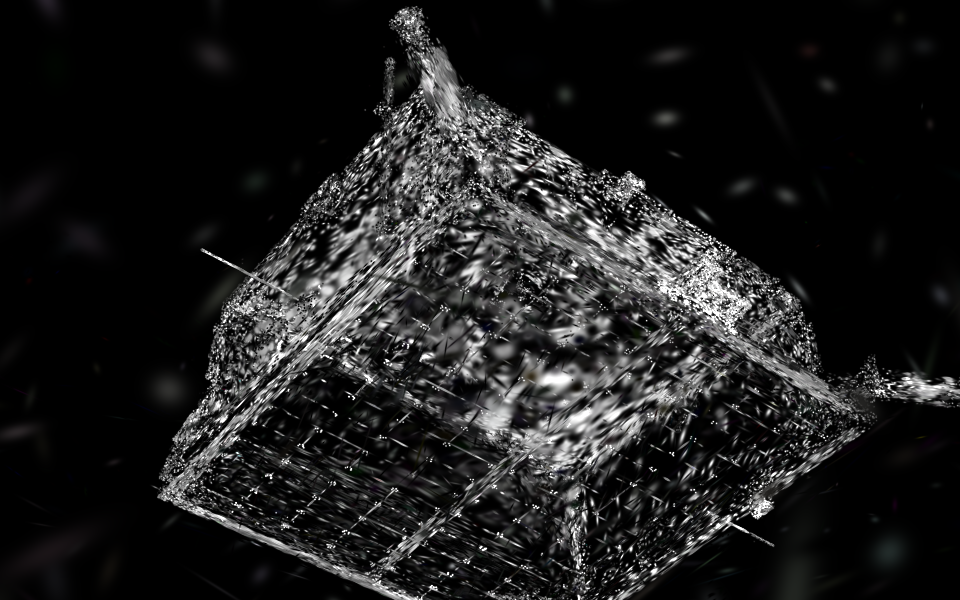}
      \caption{Gaussian splatting}
      \label{fig:gaussiansplatting}
   \end{subfigure}
   \hfill
   \begin{subfigure}[b]{0.22\textwidth}
      \centering
      \includegraphics[width=\textwidth]{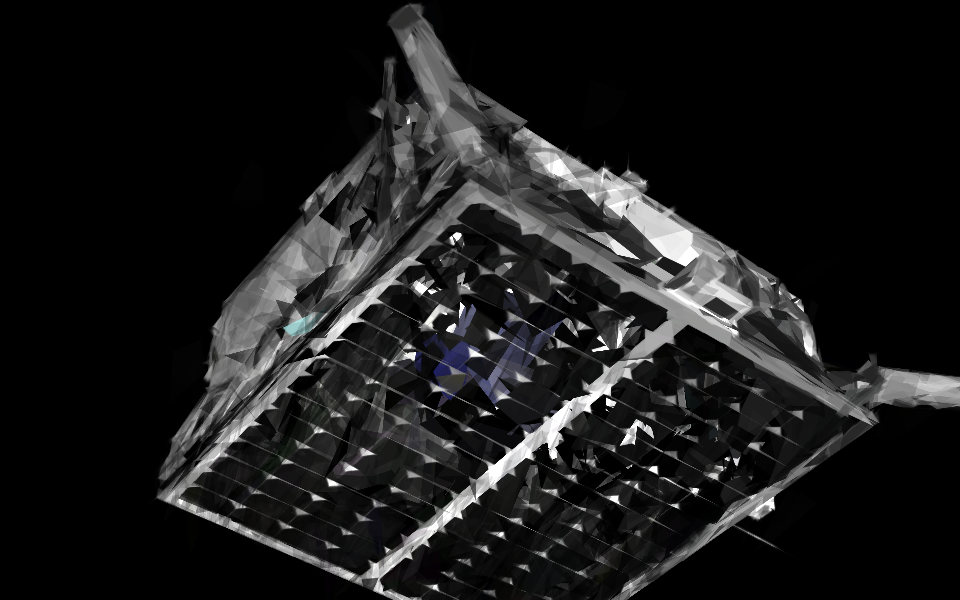}
      \caption{Convex splatting}
      \label{fig:convexsplatting}
   \end{subfigure}
   \caption{Internal explicit representation of the TANGO spacecraft. A set of primitives that represent the spacecraft is kept. The radiuses of the Gaussians were reduced and the fill ratio of the convexes were increased to make the primitives more visible.}
   \label{fig:splatting}
\end{figure}

\subsection{Preliminaries}
\label{sec:preliminaries}
We evaluate three Novel View Synthesis methods in this work: K-Planes \cite{kplanes}, Gaussian Splatting (GS) \cite{gaussiansplatting}, and Convex Splatting (CS) \cite{convexsplatting}. The primary distinction among these methods lies in their scene representation. K-Planes uses an implicit representation, encoding features that must be decoded to recover color and density of 3D points. By contrast, GS and CS are explicit representations that directly model scene geometry using 3D Gaussians and convexes, respectively. This difference is especially relevant for geometry-centric tasks, as explicit representations provide more direct control over 3D structure.

\paragraph{K-Planes.} 
K-Planes \cite{kplanes}, a variant of NeRF, represents a scene using a set of feature planes at predefined resolutions. Each 3D point is mapped to color and density values as shown in Figure~\ref{fig:kplanes}. The learnable parameters in this model are the plane features and the multilayer perceptron (MLP) weights, both initialized randomly and trained by minimizing a photometric loss between rendered and ground-truth images. The main limitation of K-Planes, and implicit methods in general, is the need to evaluate MLPs at every sampled point along a ray, which makes rendering computationally expensive. We adopt K-Planes as the baseline for implicit method due to its widespread use and its relatively efficient training and rendering compared to earlier NeRF variants.

\paragraph{Gaussian and Convex Splatting.} 
In contrast, Gaussian Splatting (GS) \cite{gaussiansplatting} and Convex Splatting (CS) \cite{convexsplatting} employ explicit scene representations, shown in Figure~\ref{fig:splatting}. A scene is modeled as a set of 3D primitives — Gaussians in GS and convexes in CS — each defined by parameters such as position, color, opacity, and size. These primitives are initialized either randomly or from a Structure-from-Motion (SfM) point cloud \cite{sfm_survey}, and are rendered by projecting them onto the image plane and accumulating their contributions to each pixel. Training again minimizes a photometric loss, but directly updates the primitive parameters rather than MLP weights. Explicit representations have two main advantages: they enable highly parallel rendering without expensive per-sample MLP evaluations, and they offer more direct control over geometric structure.

A common failure mode of the original formulations of these methods is sensitivity to illumination changes, which can introduce artifacts in rendered images. In space-based scenarios, such variations arise naturally from changes in the relative position of the Sun with respect to the target object. To address this challenge, appearance embeddings were introduced in NeRF in the Wild \cite{nerfinthewild}, providing additional flexibility in color prediction. These embeddings are optimized jointly with the scene parameters, encode per-image lighting conditions, and are incorporated as an extra input when computing the color of a point or primitive.

Extensions of this appearance embedding idea already exist for K-Planes \cite{kplanes} and Gaussian Splatting, where it is referred to as \emph{WildGaussians} \cite{wildgaussians}. In this work, we extend the approach to Convex Splatting, introducing \emph{WildConvexes}. We refer to the original methods without appearance embeddings as \emph{vanilla} K-Planes, GS, and CS. While appearance embeddings were originally designed to improve robustness to lighting, our study examines whether they also mitigate floater artifacts in geometry-critical reconstruction tasks.

\subsection{Experimental protocol}
\label{sec:experiments}
\paragraph{Dataset}
Our experiments are conducted on the SPEED+ dataset \cite{speedplus}, which provides synthetic and hardware-in-the-loop imagery of the TANGO spacecraft \cite{prismamission} together with ground-truth pose labels. The dataset captures the spacecraft under varying poses and illumination conditions, making it a representative benchmark for space robotics. For our experiments, we use 100 synthetic training images without Earth in the background and 20 synthetic test images, downscaled by a factor of two to a resolution of 960x600.

\paragraph{Metrics}
Our primary objective is to assess the geometric fidelity of the learned representations. Ideally, this evaluation would be conducted using three-dimensional metrics. However, the SPEED+ dataset lacks ground-truth 3D geometry, which constitutes a significant limitation. Consequently, we evaluate the performance of different methods in image space by comparing thresholded depth maps rendered from each approach against the ground-truth masks provided in the dataset. The thresholded depth maps were generated by min-max normalizing the depth maps and thresholding these at $0.5$.

The metric used in prior space-focused NVS work \cite{nerfgrafspace} is Intersection over Union (IoU), which measures the overall alignment between predicted and ground-truth masks. While informative, IoU alone does not distinguish between different sources of error. To obtain a more complete picture, we report three additional metrics, all computed from the counts of true positives (tp), false positives (fp), true negatives (tn), and false negatives (fn), averaged across the 20 test images:
\begin{itemize}
   \item \emph{True positive rate (TPR)}: $\frac{\textnormal{tp}}{\textnormal{tp}+\textnormal{fn}}$, with optimal value 1. This quantifies how much of the ground-truth object is correctly captured, relevant in space robotics as low TPR increases collision risk.
   \item \emph{False positive rate (FPR)}: $\frac{\textnormal{fp}}{\textnormal{fp}+\textnormal{tn}}$, with optimal value 0. This reflects the amount of empty space erroneously filled by geometry, typically manifesting as floaters.
   \item \emph{False discovery rate (FDR)}: $\frac{\textnormal{fp}}{\textnormal{fp}+\textnormal{tp}}$, with optimal value 0. This indicates the fraction of predicted geometry that is spurious, directly tied to the trustworthiness of the representation. Useful for contact tasks.
\end{itemize}

We additionally evaluate the efficiency of each representation by reporting the number of learned parameters. For K-Planes, this includes the features stored in the planes and the MLP weights. For explicit methods, it equals the number of primitives multiplied by their parameters. This metric enables a fair comparison across implicit and explicit methods, and better reflects onboard feasibility than simply reporting primitive counts, since convexes require substantially more parameters than Gaussians. Appearance embedding parameters are excluded, as they are only used during training and do not affect the converged geometry, nor need to be transmitted when sharing it. Their contribution to the parameter count is minor in GS ($\sim$1\% of total parameters), but more significant in CS ($\sim$25\%).

Finally, although our focus is geometric, we also report standard photometric metrics — PSNR, SSIM, and LPIPS — as auxiliary indicators. These scores degrade when geometric artifacts appear, and qualitative inspection of renderings further highlights differences between methods and exposes failure cases not captured numerically.

\paragraph{Appearance Embeddings}
Accurate photometric evaluation necessitates rendering under lighting conditions that correspond to those present in the ground-truth images. In \emph{in the wild} models, these conditions can be regulated through appearance embeddings. Existing studies \cite{nerfinthewild, wildgaussians} optimize these appearance embeddings during evaluation, prior to computing photometric performance metrics. However, this optimization requires access to camera poses and ground-truth images, which constitutes privileged information that artificially enhances model performance. To ensure equitable comparison across all methods, we abstain from this fine-tuning procedure and instead employ randomly initialized appearance embeddings during evaluation. We found this methodological choice produces negligible effects on geometry-based metrics while influencing only photometric scores.

\paragraph{Initialization}
To ensure fairness, all models are initialized randomly. K-Planes requires random initialization, as its plane features and MLPs cannot easily incorporate prior knowledge. Explicit methods (GS/CS) could in principle be initialized from a point cloud reconstructed with structure-from-motion (SfM) methods such as COLMAP \cite{colmap, colmap2}, but in practice these struggle in the featureless backgrounds typical of space imagery and offer little benefit when they do converge. All models are trained for 60,000 iterations, consistent with prior work \cite{convexsplatting, gaussiansplatting, wildgaussians}.

\section{Results}
Given the dataset, metrics, and training setup defined in Section~\ref{sec:experiments}, we now evaluate the three methods under identical conditions and report both quantitative and qualitative results.

\label{sec:results}
\begin{table*}
   \centering
   \caption{Geometric metrics averaged over 20 training images after 60k training iterations. \colorbox{firstcolor}{First}, \colorbox{secondcolor}{second} and \colorbox{thirdcolor}{third} best results are highlighted. For the splatting models, the number of primitives is added in parentheses.}
   \begin{tabular}{c|c|c|c|c|c}
      \toprule
      \textbf{Model} & \textbf{IoU $\uparrow$} & \textbf{TPR $\uparrow$} & \textbf{FPR $\downarrow$} & \textbf{FDR $\downarrow$} & \textbf{\# Parameters $\downarrow$} \\
      \midrule
      \textbf{VKP} & 0.18 & \cellcolor{thirdcolor}0.93 & 0.81 & 0.81 & 33 935 378 \\
      \textbf{WKP} & 0.17 & 0.87 & 0.86 & 0.83 & 33 940 626 \\
      \textbf{VGS} & \cellcolor{secondcolor}0.48 & 0.90 & \cellcolor{secondcolor}0.15 & \cellcolor{secondcolor}0.47 & 4 298 150 (69 325) \\
      \textbf{WGS} & \cellcolor{firstcolor}0.53 & \cellcolor{secondcolor}0.96 & \cellcolor{firstcolor}0.13 & \cellcolor{firstcolor}0.45 & \cellcolor{thirdcolor}2 113 574 (23 748) \\
      \textbf{VCS} & 0.37 & \cellcolor{firstcolor}1.00 & 0.30 & 0.63 & \cellcolor{secondcolor}461 127 (6 683) \\
      \textbf{WCS} & \cellcolor{thirdcolor}0.40 & \cellcolor{firstcolor}1.00 & \cellcolor{thirdcolor}0.26 & \cellcolor{thirdcolor}0.60 & \cellcolor{firstcolor}418 002 (6 058) \\
      \bottomrule
   \end{tabular}
   \label{tab:geometric}
\end{table*}

\subsection{Geometric Performance}
Table~\ref{tab:geometric} shows the geometric performance of the different models we introduced. Inspection of these results reveals a clear distinction between implicit and explicit methods across all metrics. Both variants of K-Planes perform the worst, requiring significantly more parameters while yielding poor IoU, TPR, and especially FPR and FDR. The high FDR illustrates the abundance of density occupying empty space around the spacecraft. This is corroborated by the depth renderings in Table~\ref{tab:qualitative}, where the spacecraft is barely distinguishable and most geometry is misplaced.

In contrast, the explicit methods achieve higher IoU and nearly perfect TPR, meaning the geometry of the ground-truth spacecraft is consistently recovered. They also require substantially fewer parameters, highlighting their efficiency in resource-constrained settings such as spacecraft. A striking difference emerges between GS and CS: convex splatting achieves comparable quality with far fewer primitives, underscoring its efficiency, though at the cost of a higher FPR. Qualitative inspection suggests this is due to convex primitives extending beyond the true spacecraft surface, making convexes well-suited for large, regular structures but less precise at boundaries. This efficiency-quality trade-off could be advantageous for downstream interaction tasks, where compact, part-like decompositions are beneficial.

Within each explicit method, the addition of appearance embeddings brings a modest but consistent geometric improvement. This effect stems from their ability to explain illumination changes during training, even though embeddings have no influence at inference. For GS, embeddings also reduce the number of required primitives, showing their positive impact on parameter efficiency. For CS, embeddings do not affect parameter count, but they still provide minor geometric benefits.

\begin{table}
   \centering
   \caption{Photometric metrics averaged over 20 training images after 60k training iterations.\colorbox{firstcolor}{First}, \colorbox{secondcolor}{second} and \colorbox{thirdcolor}{third} best results are highlighted. * corresponds to appearance embeddings trained on the ground truth image at evaluation, no * corresponds to random appearance embeddings.}
   \begin{tabular}{c|c|c|c}
      \toprule
      \textbf{Model} & \textbf{PSNR $\uparrow$} & \textbf{SSIM $\uparrow$} & \textbf{LPIPS $\downarrow$}\\
      \midrule
      \textbf{VKP} & 19.85 & \cellcolor{thirdcolor}0.82 & \cellcolor{secondcolor}0.41 \\
      \textbf{WKP} & 18.63 & 0.52 & \cellcolor{thirdcolor}0.46 \\
      \textbf{WKP*} & 20.70 & 0.55 & \cellcolor{thirdcolor}0.46 \\
      \textbf{VGS} & 22.23 & \cellcolor{secondcolor}0.84 & 0.47 \\
      \textbf{WGS} & 20.86 & 0.77 & \cellcolor{firstcolor}0.38 \\
      \textbf{WGS*} & \cellcolor{firstcolor}25.22 & 0.79 & \cellcolor{firstcolor}0.38 \\
      \textbf{VCS} & \cellcolor{thirdcolor}22.50 & \cellcolor{firstcolor}0.85 & 0.48 \\
      \textbf{WCS} & 21.75 & \cellcolor{secondcolor}0.84 & 0.49 \\
      \textbf{WCS*} & \cellcolor{secondcolor}23.06 & \cellcolor{firstcolor}0.85 & 0.49 \\
      \bottomrule
   \end{tabular}
   \label{tab:photometric}
\end{table}

\subsection{Photometric Performance}
The photometric metrics reported in Table~\ref{tab:photometric} reinforce the trends observed in geometry. K-Planes again performs the worst, while GS and CS achieve comparable performance, confirming the advantage of explicit representations over implicit ones.  

The effect of appearance embeddings is more pronounced here, especially in PSNR. Random embeddings sampled at inference degrade performance relative to the \emph{vanilla} models, as they typically correspond to lighting conditions less consistent than the average appearance learned during training. Conversely, when appearance embeddings are optimized at inference time using the ground-truth image, PSNR improves substantially beyond the \emph{vanilla} baseline. This highlights the role of embeddings in capturing illumination, though at the cost of introducing leakage from the reference image.  

Interestingly, SSIM and LPIPS are less sensitive to embeddings. Unlike PSNR, which directly reflects pixel-wise fidelity, these metrics emphasize structural similarity and perceptual consistency, making them less dependent on color or lighting variations.

\begin{table*}[h]
   \centering
   \caption{Qualitative results on test images from the SPEED+ dataset. The first row shows the ground truth and rendered image from each representation, while the second row shows the corresponding depth maps. * corresponds to appearance embeddings trained on the ground truth image, without corresponds to random appearance embeddings.}
   \begin{tabular}{@{}c@{}|@{}c@{} @{}c@{} @{}c@{} @{}c@{} @{}c@{} @{}c@{} @{}c@{} @{}c@{} @{}c@{}}
      \toprule
      Ground truth & VKP & WKP & WKP* & VGS & WGS & WGS* & VCS & WCS & WCS*\\
      \midrule
      \includegraphics[width=0.1\textwidth]{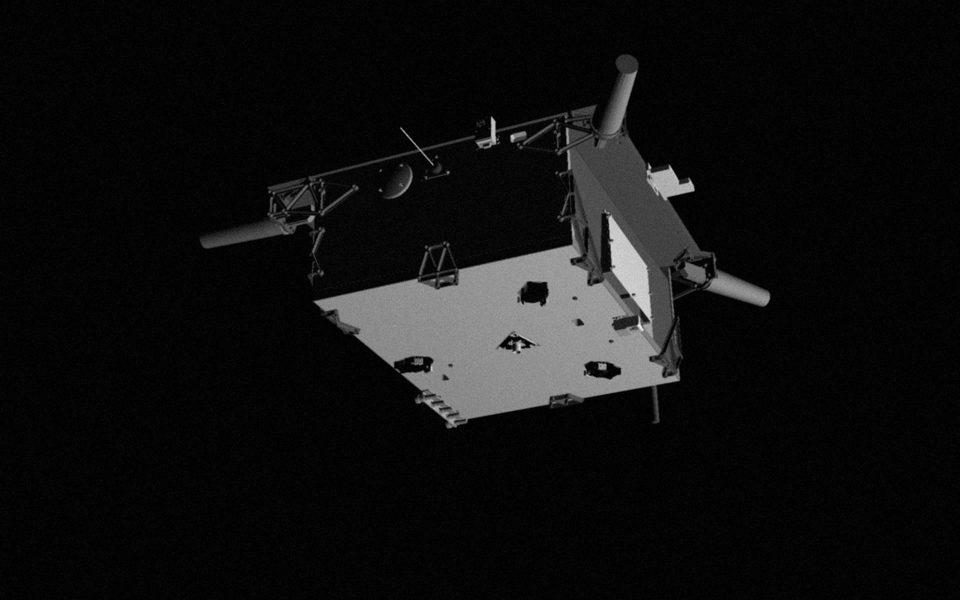} & \includegraphics[width=0.1\textwidth]{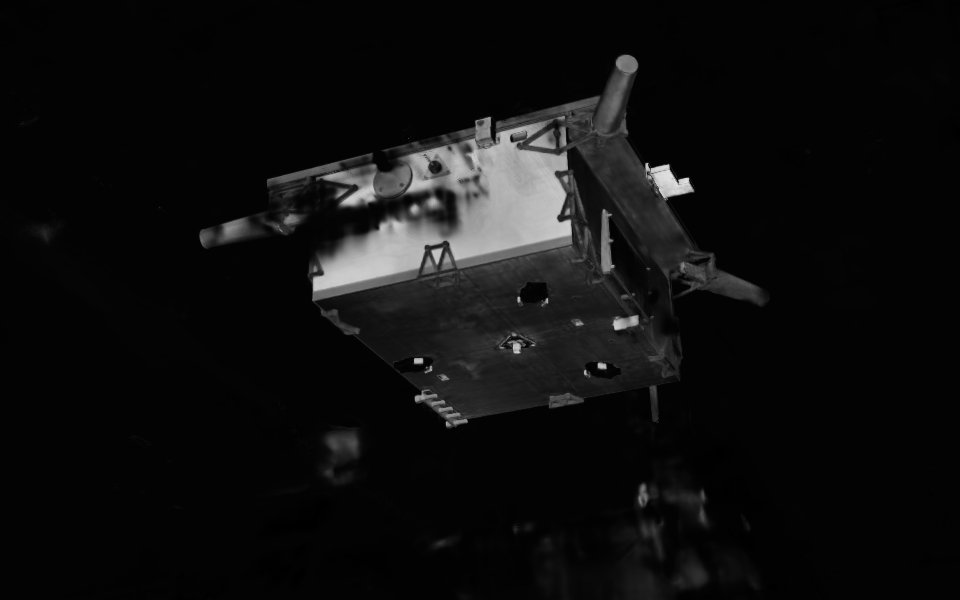} & \includegraphics[width=0.1\textwidth]{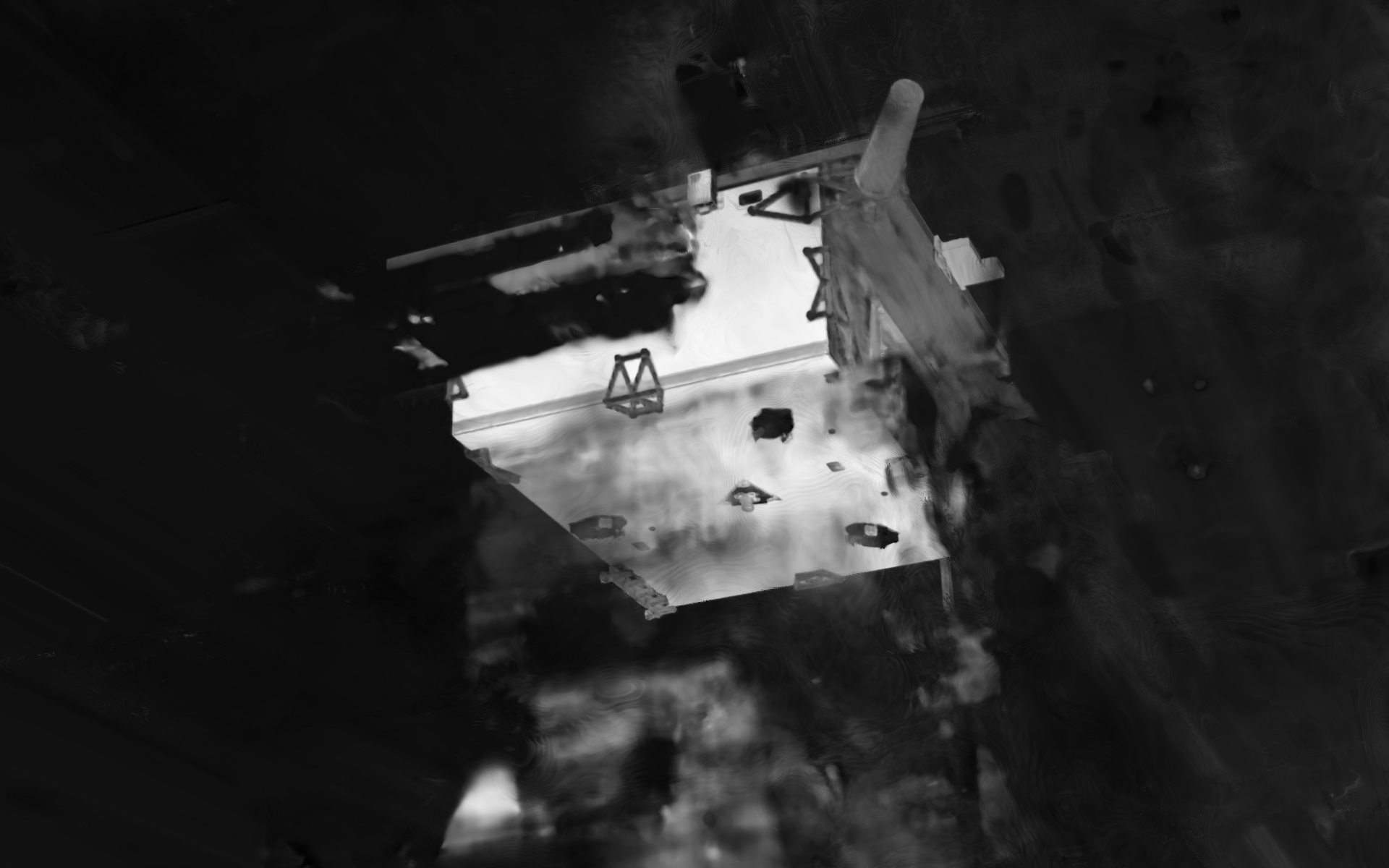} & \includegraphics[width=0.1\textwidth]{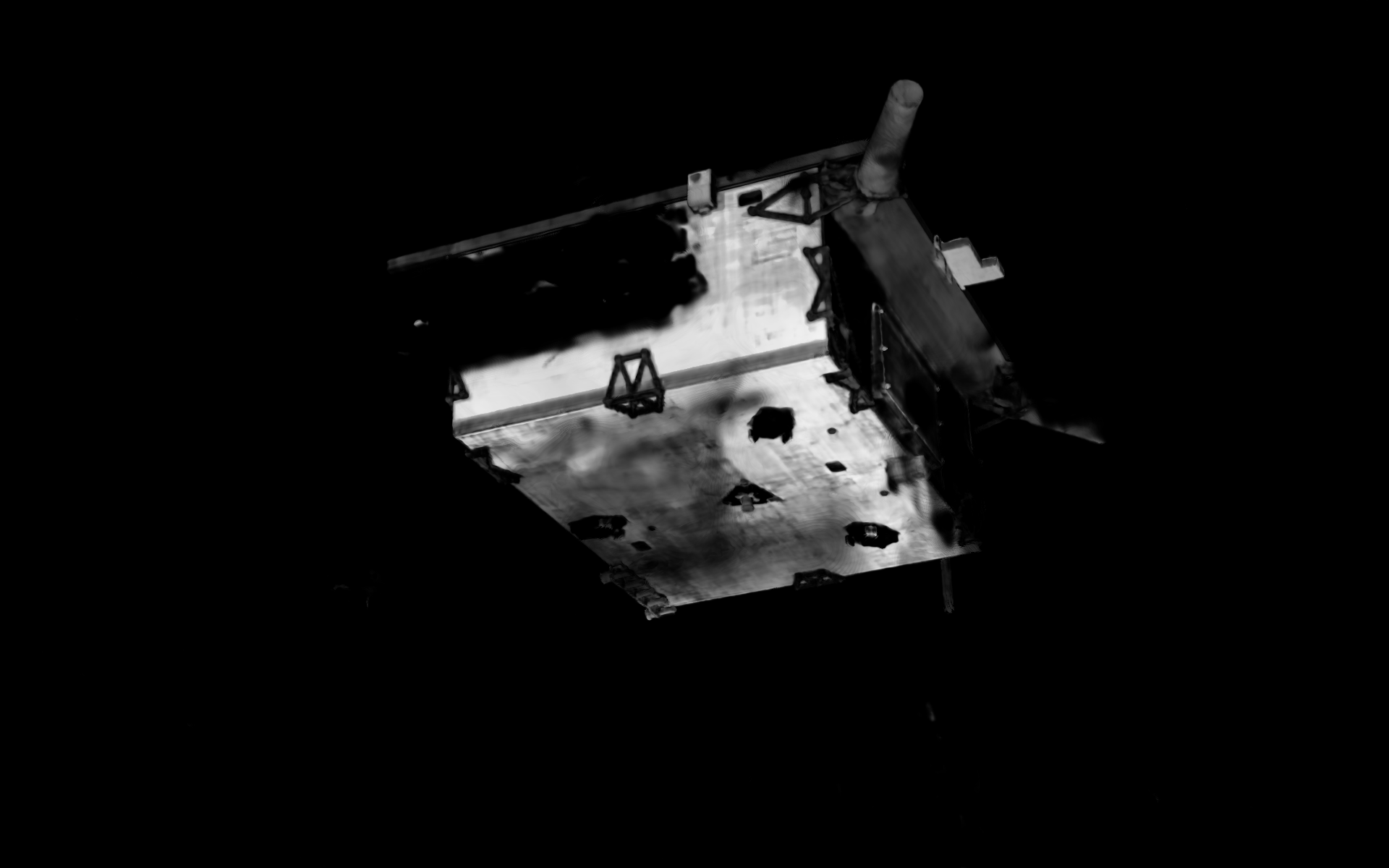} & \includegraphics[width=0.1\textwidth]{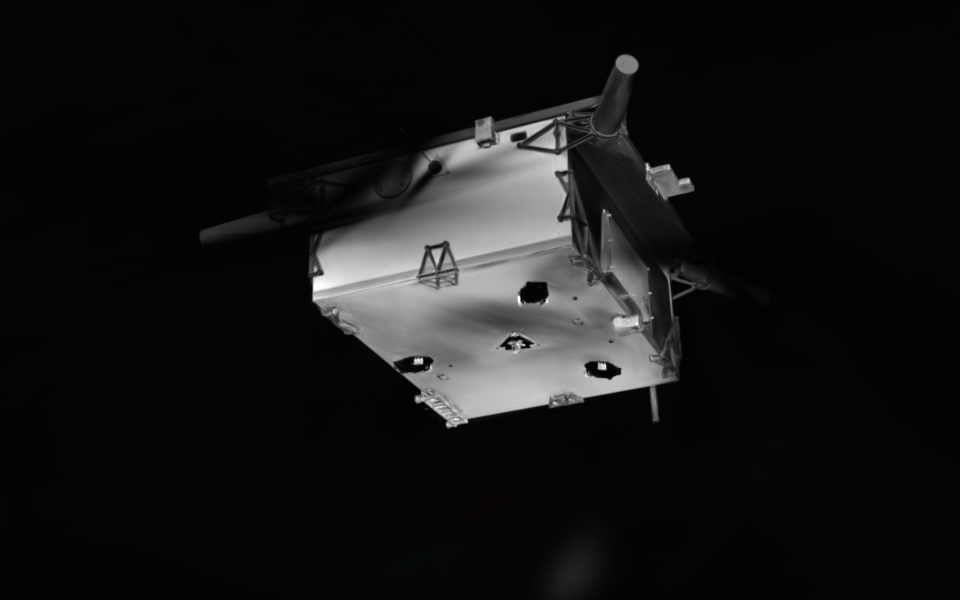} & \includegraphics[width=0.1\textwidth]{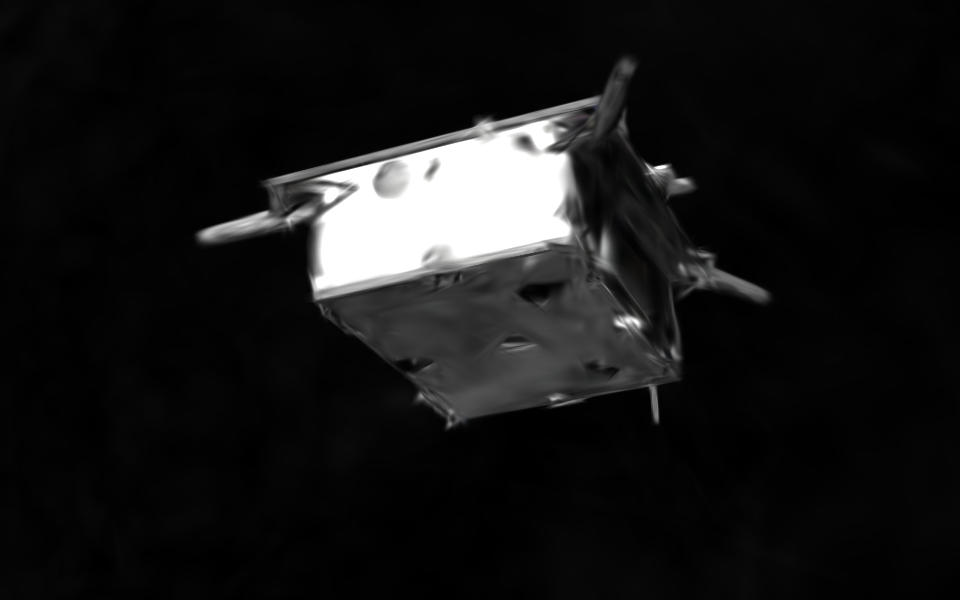} & \includegraphics[width=0.1\textwidth]{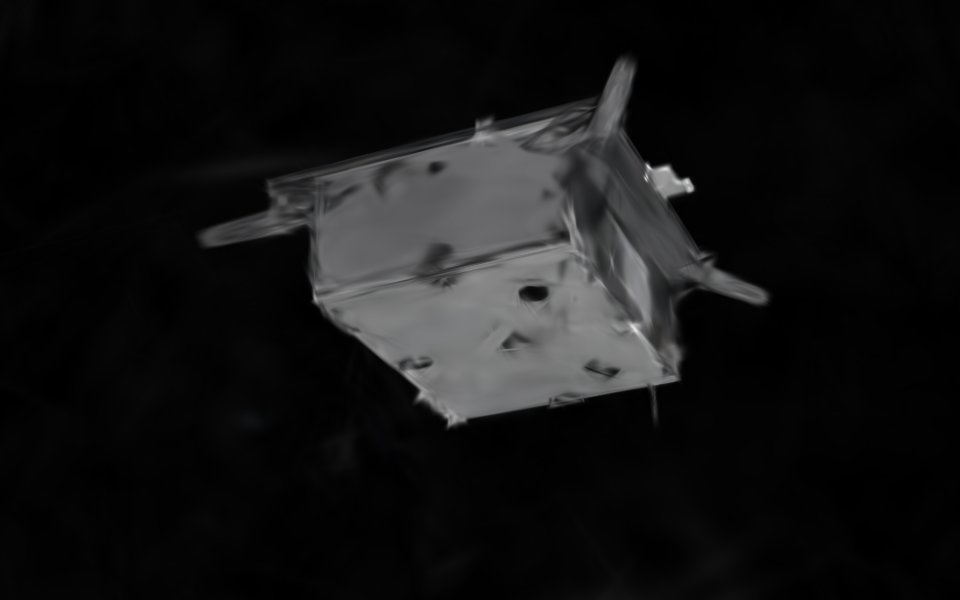} & \includegraphics[width=0.1\textwidth]{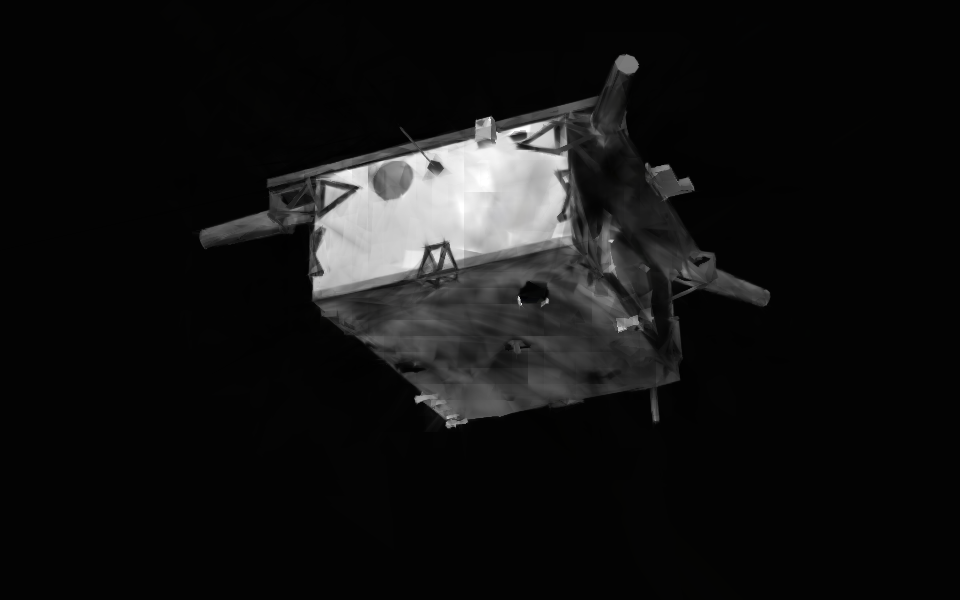} & \includegraphics[width=0.1\textwidth]{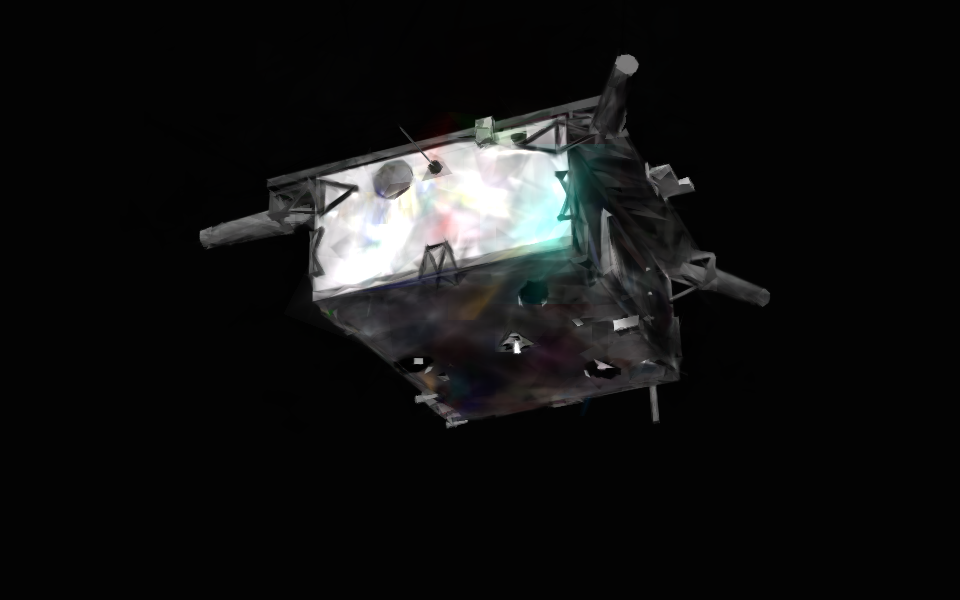} & \includegraphics[width=0.1\textwidth]{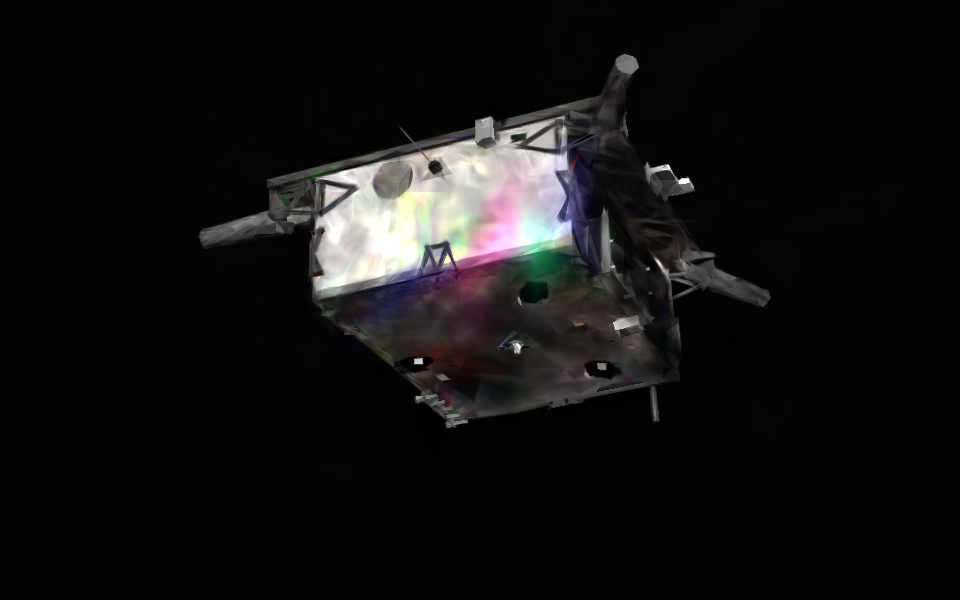} \\
      & \includegraphics[width=0.1\textwidth]{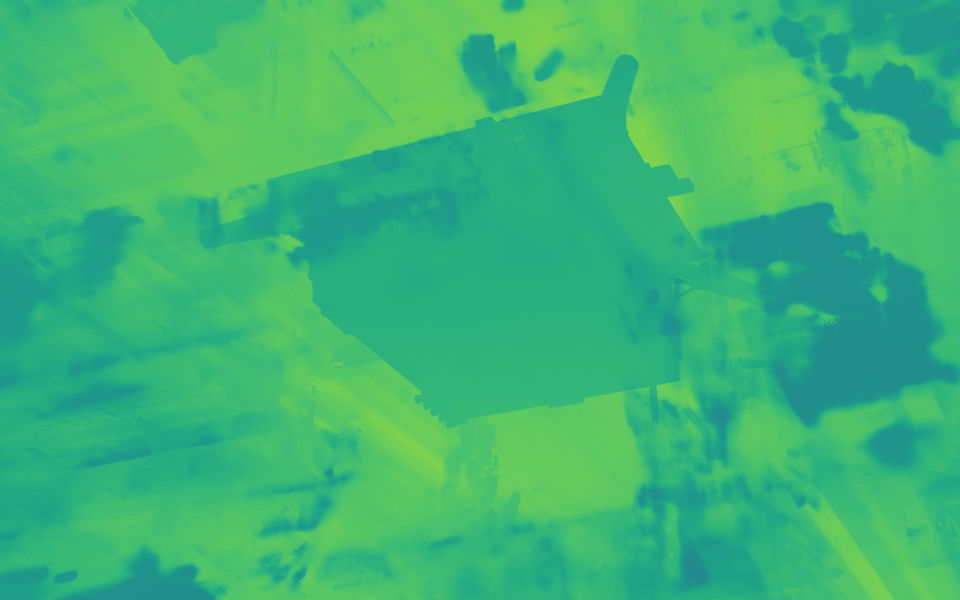} & \includegraphics[width=0.1\textwidth]{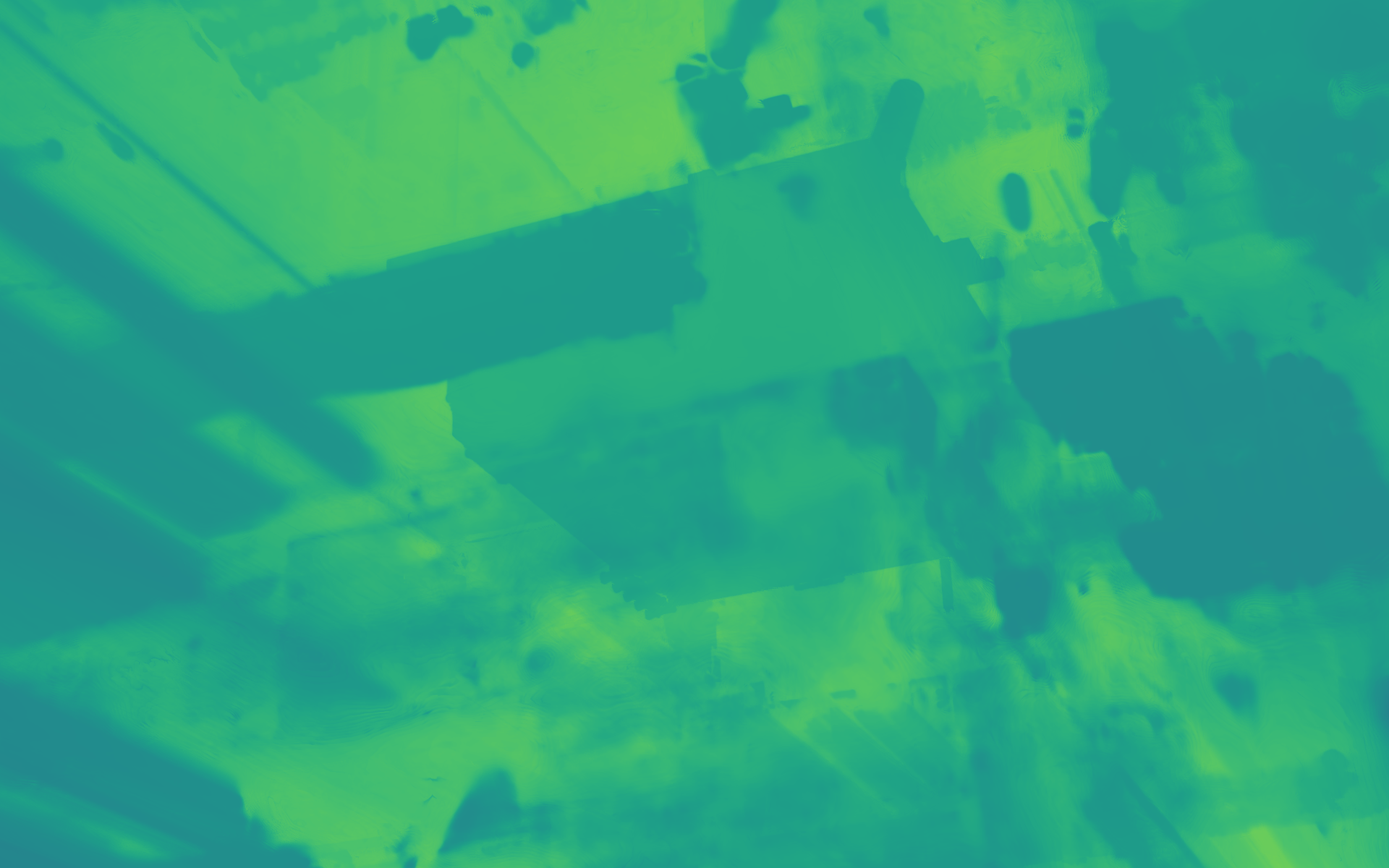} & \includegraphics[width=0.1\textwidth]{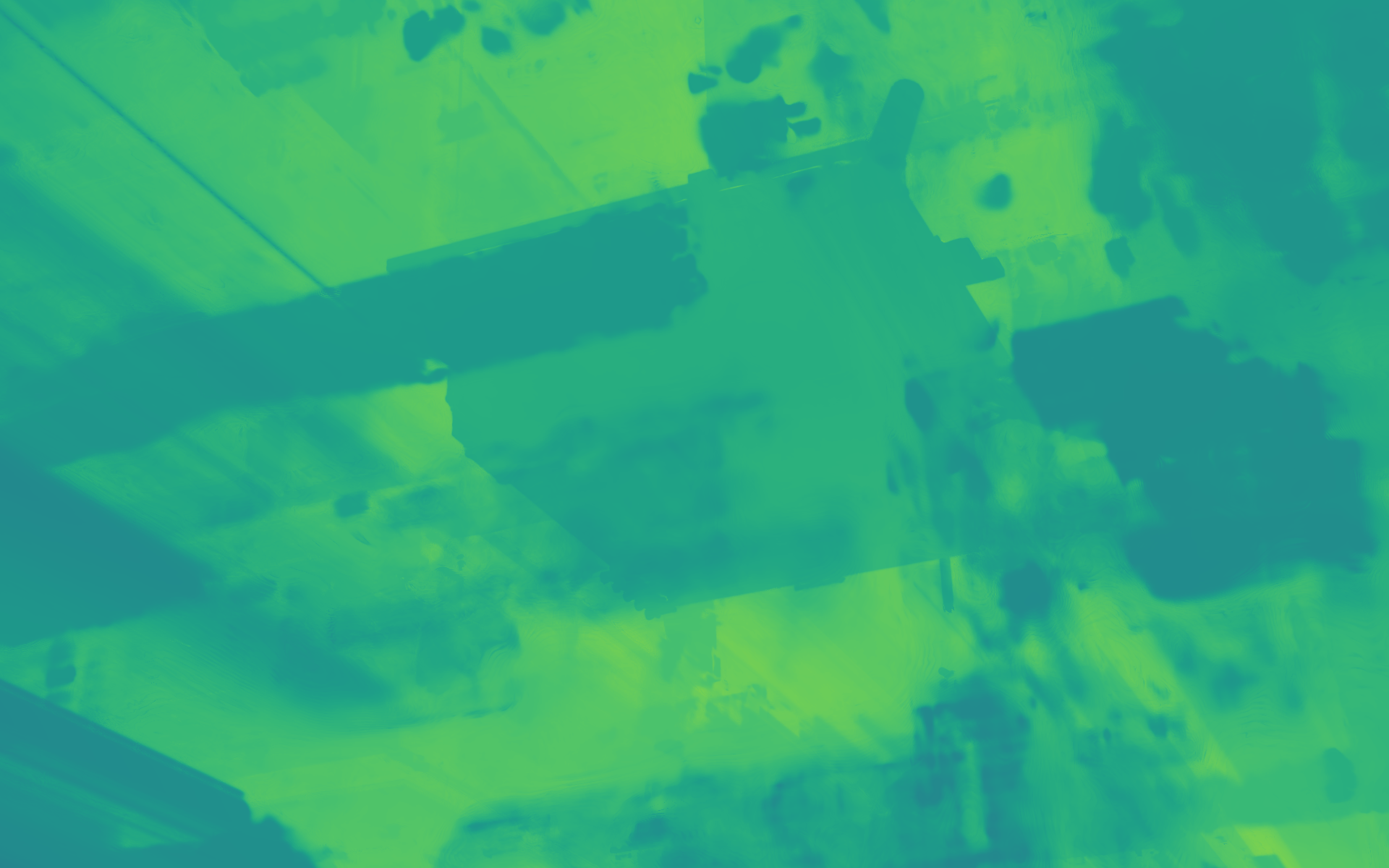} & \includegraphics[width=0.1\textwidth]{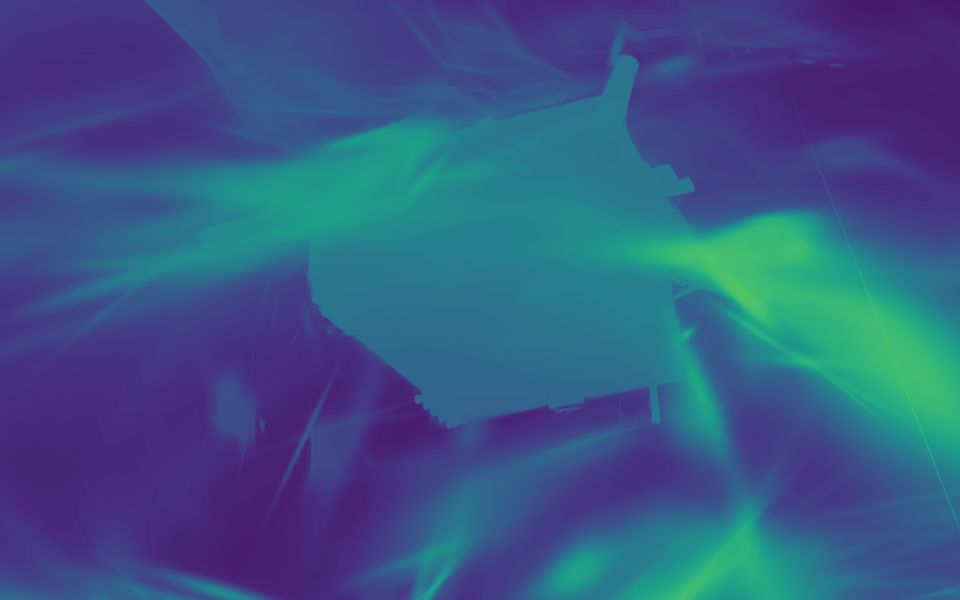} & \includegraphics[width=0.1\textwidth]{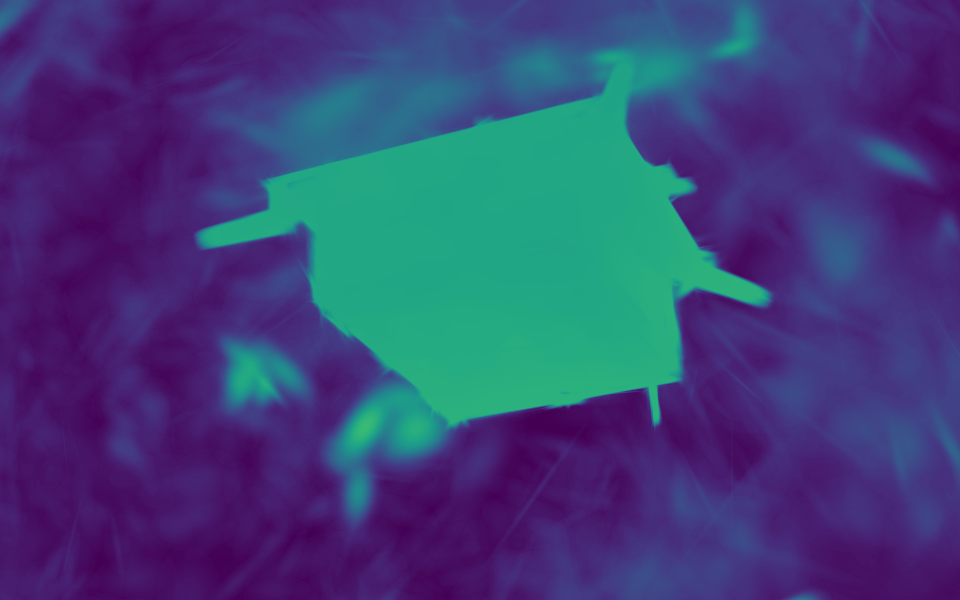} & \includegraphics[width=0.1\textwidth]{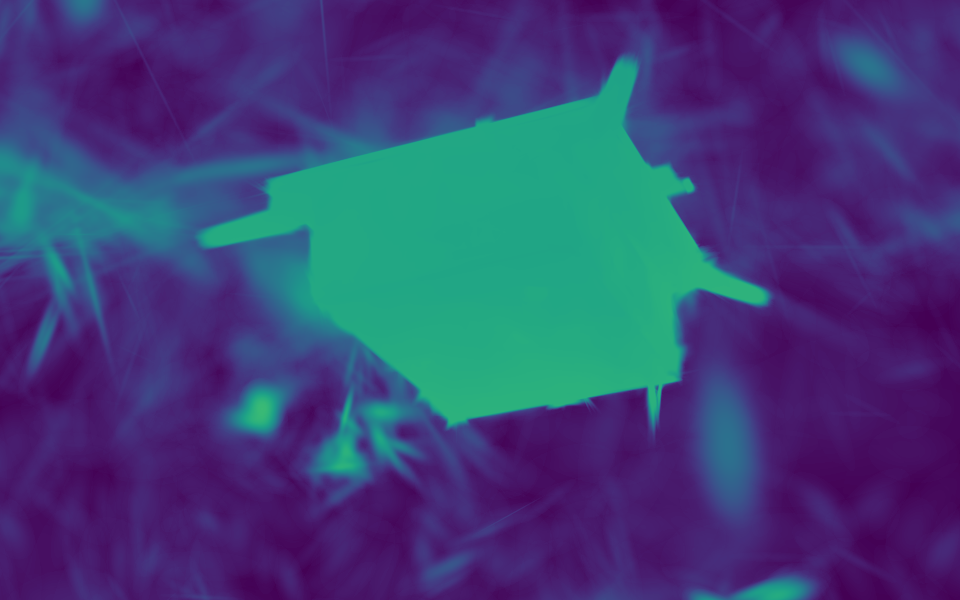} & \includegraphics[width=0.1\textwidth]{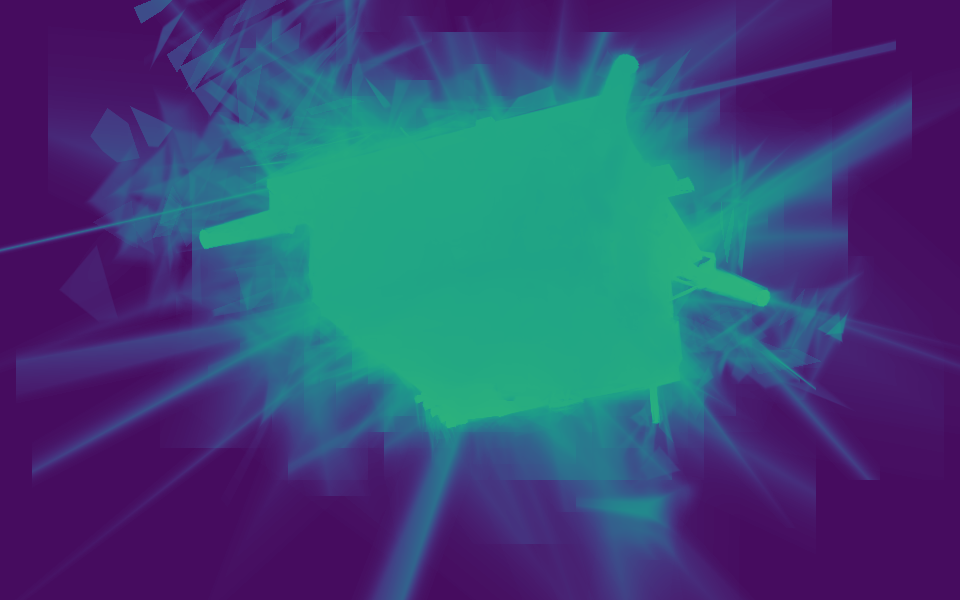} & \includegraphics[width=0.1\textwidth]{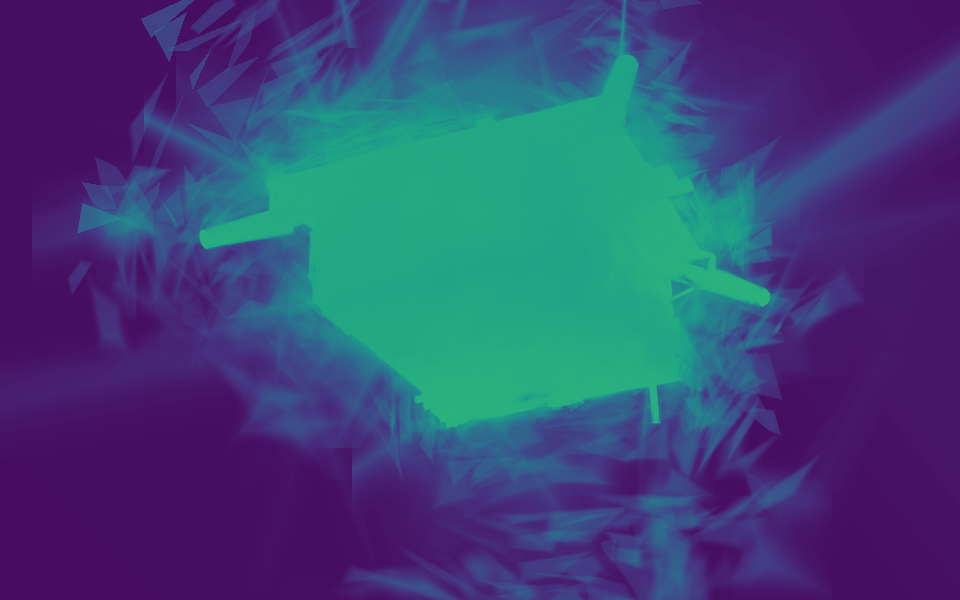} & \includegraphics[width=0.1\textwidth]{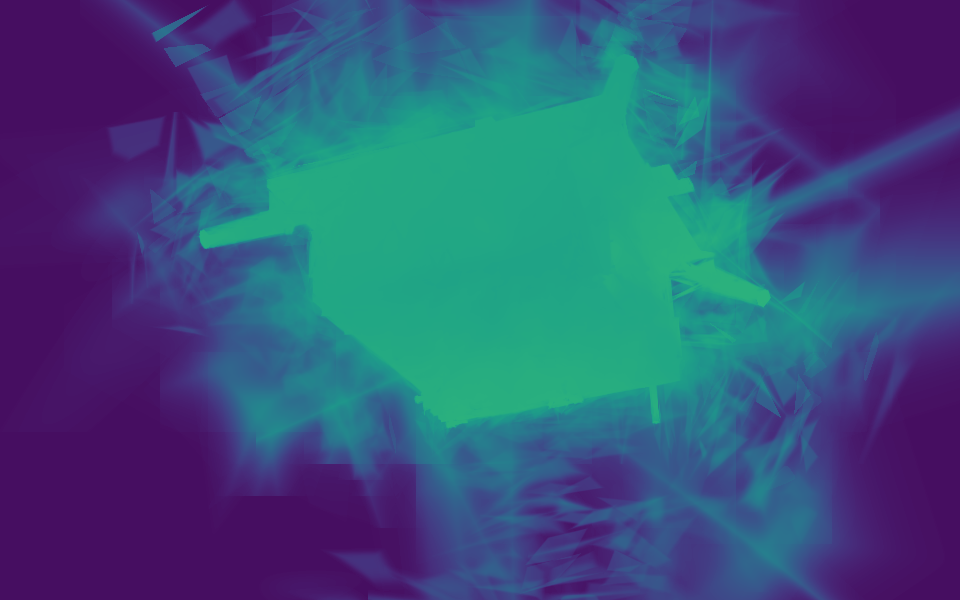} \\
      \midrule
      \includegraphics[width=0.1\textwidth]{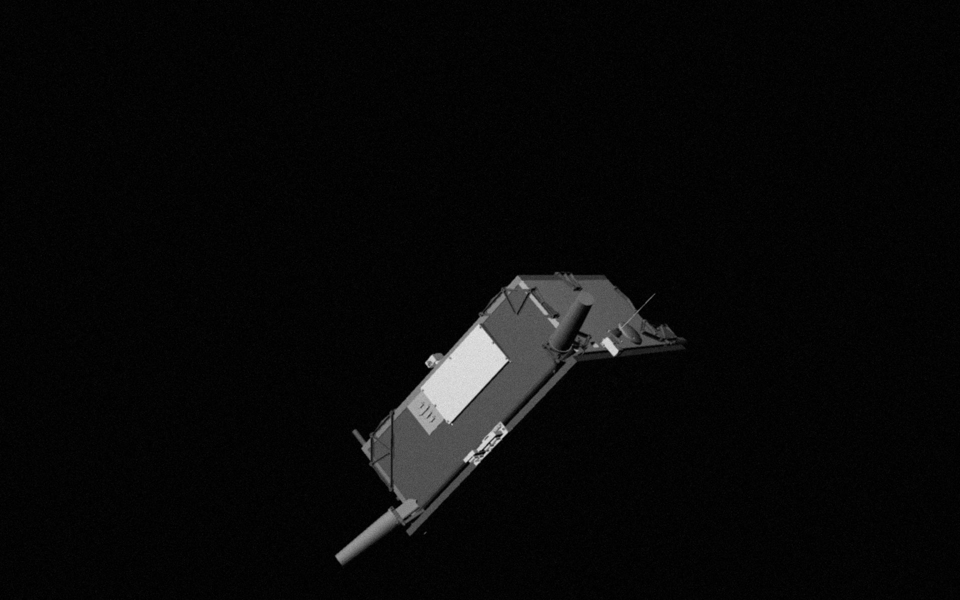} & \includegraphics[width=0.1\textwidth]{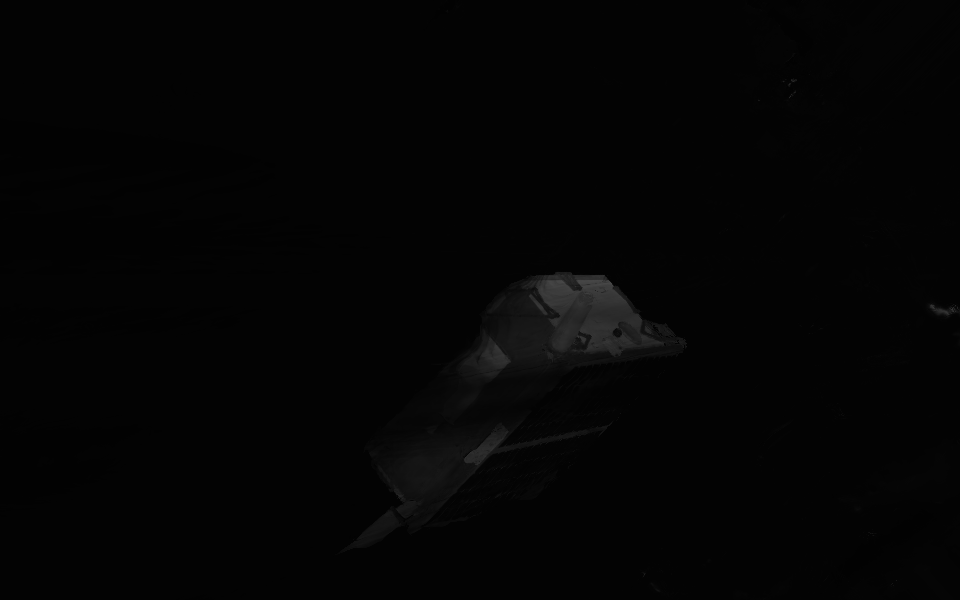} & \includegraphics[width=0.1\textwidth]{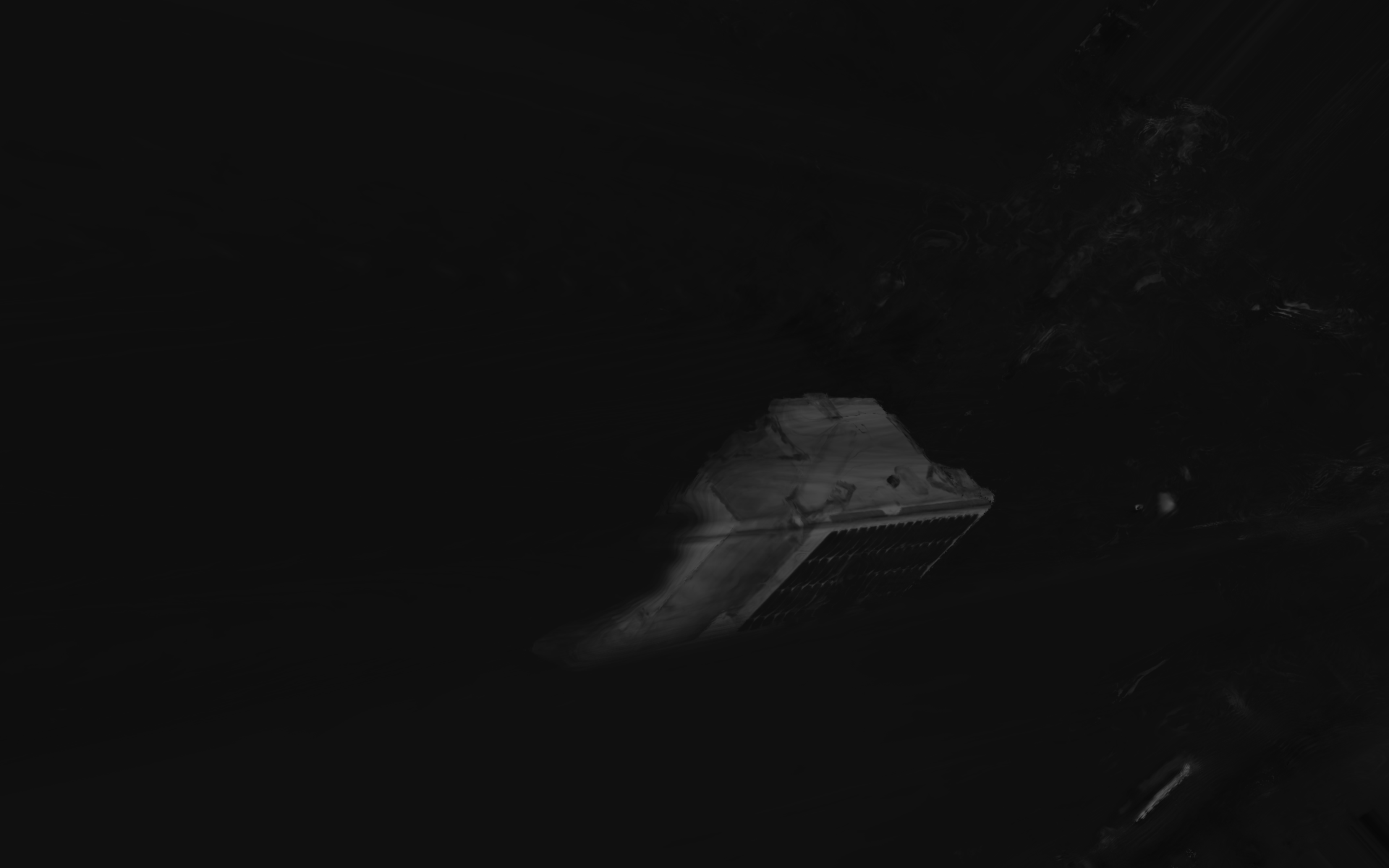} & \includegraphics[width=0.1\textwidth]{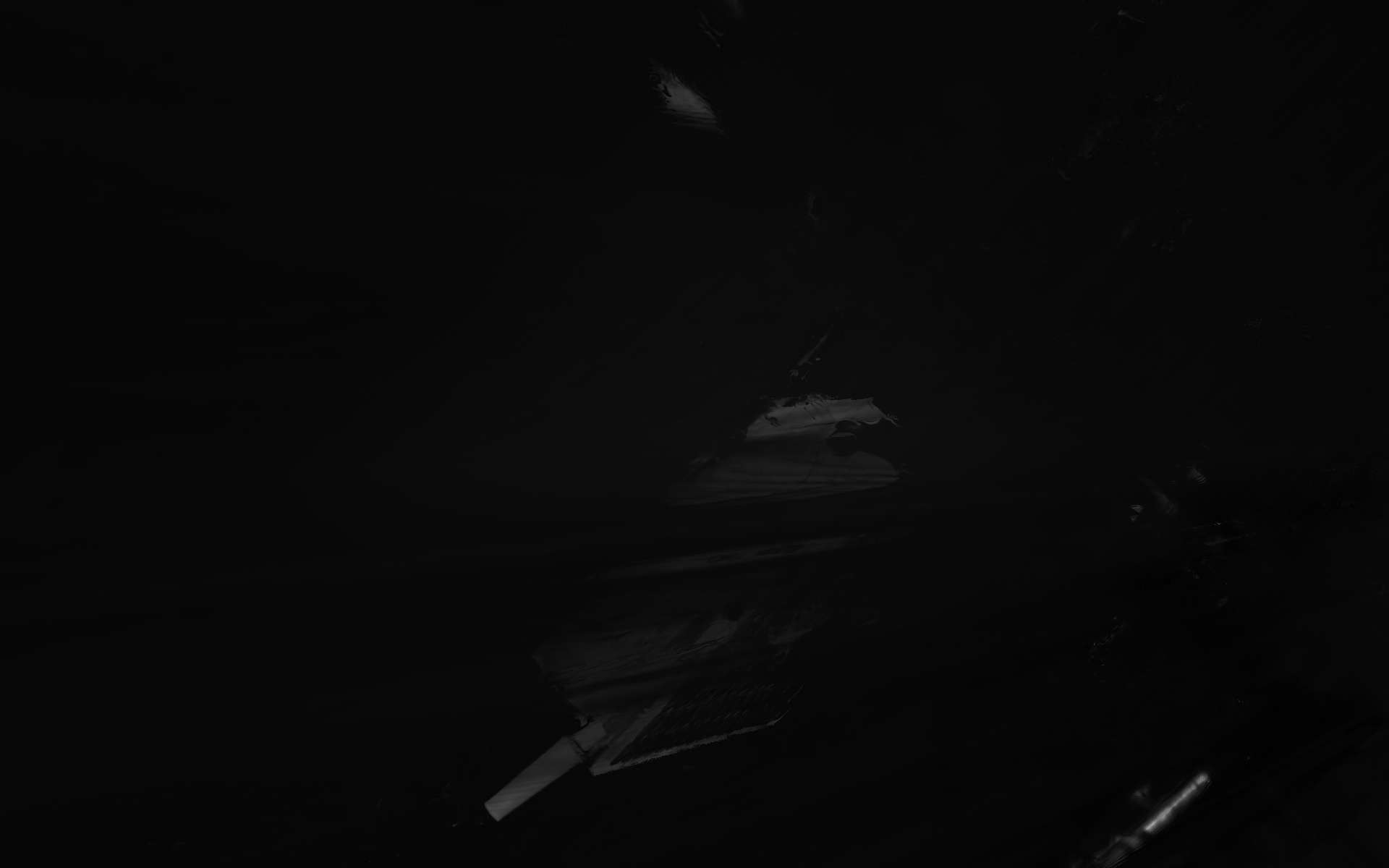} & \includegraphics[width=0.1\textwidth]{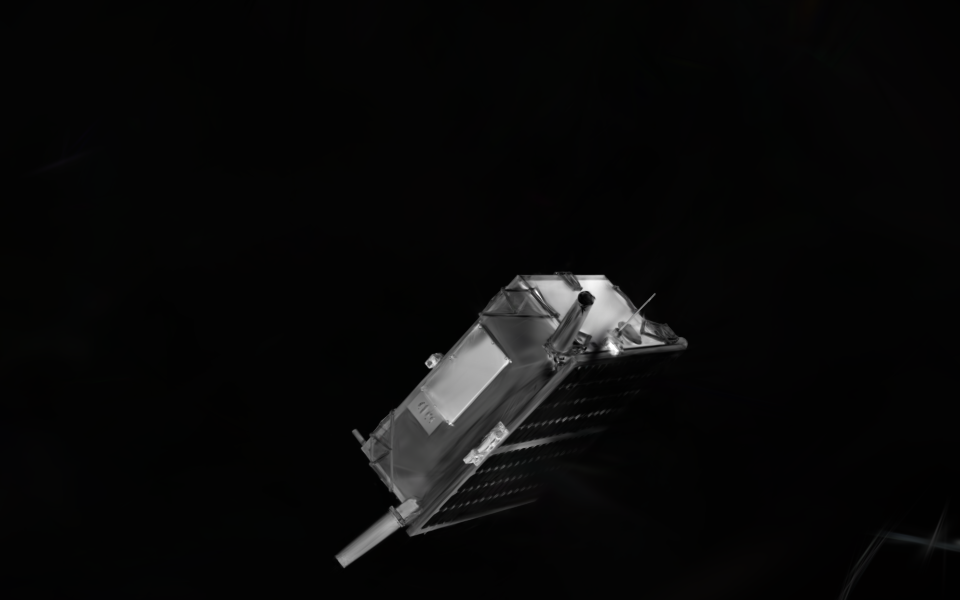} & \includegraphics[width=0.1\textwidth]{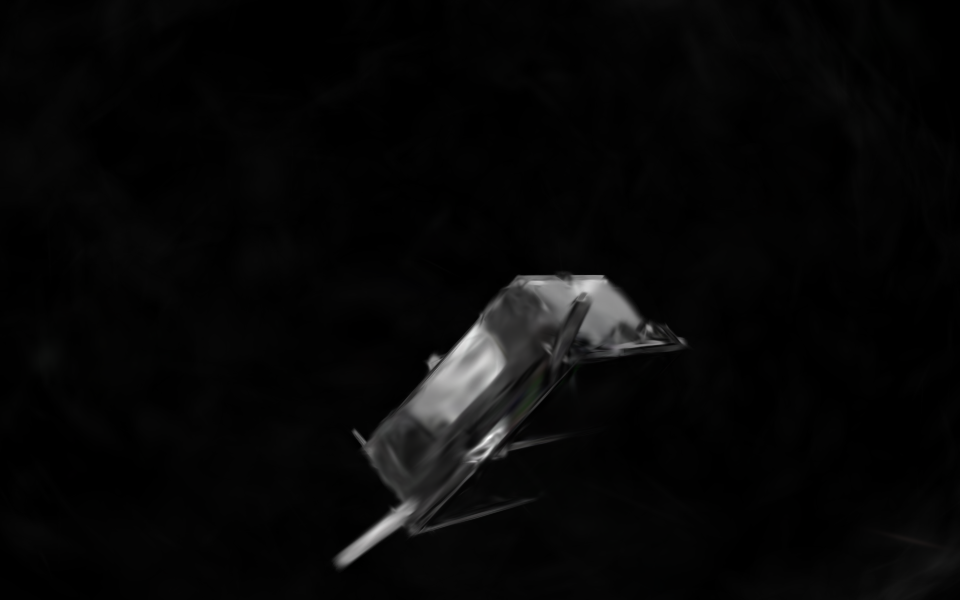} & \includegraphics[width=0.1\textwidth]{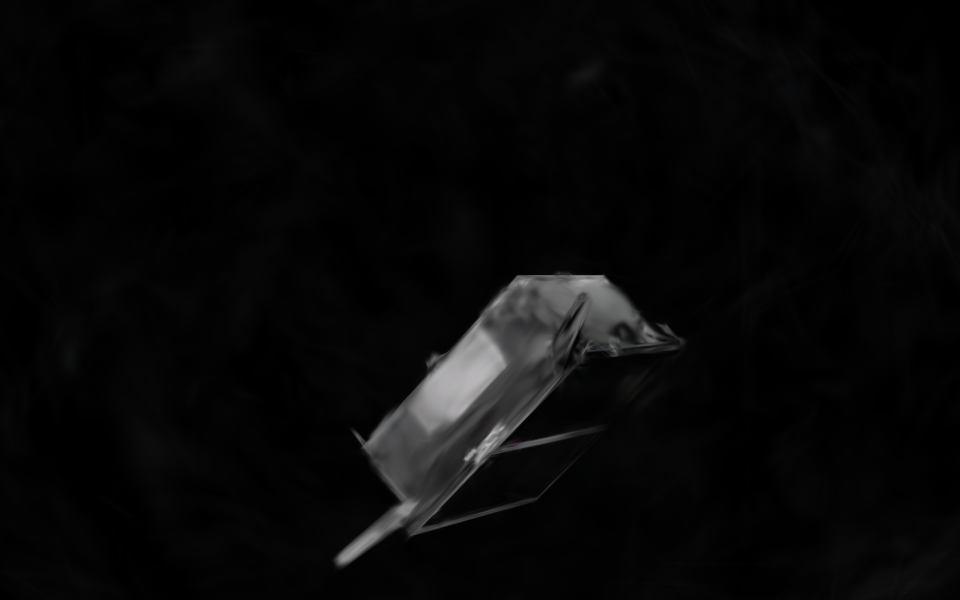} & \includegraphics[width=0.1\textwidth]{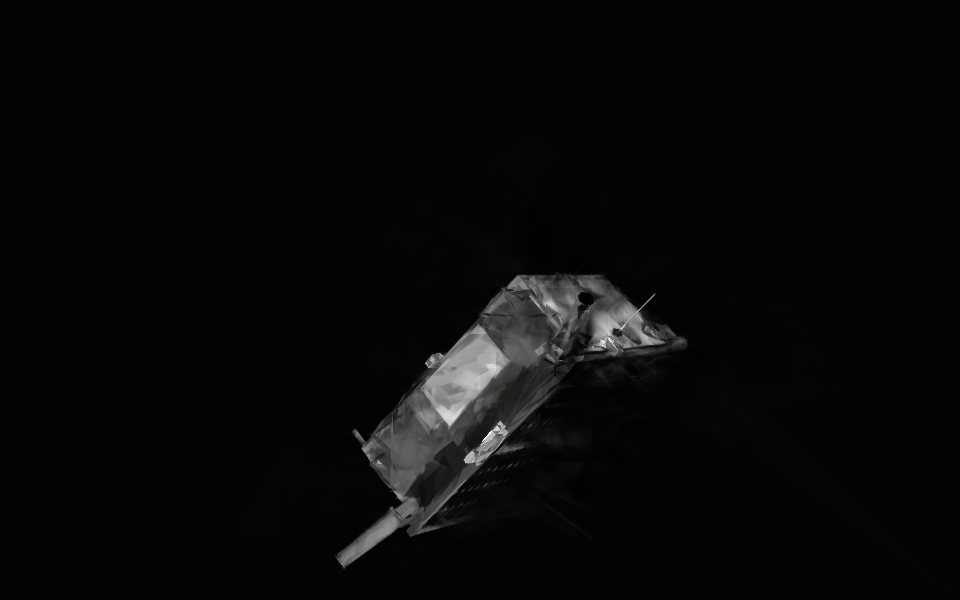} & \includegraphics[width=0.1\textwidth]{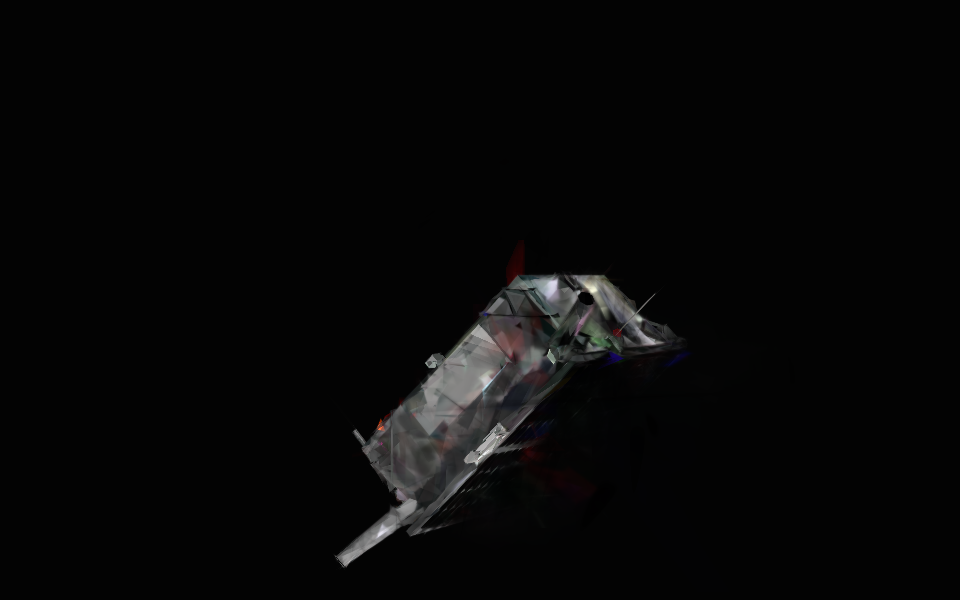} & \includegraphics[width=0.1\textwidth]{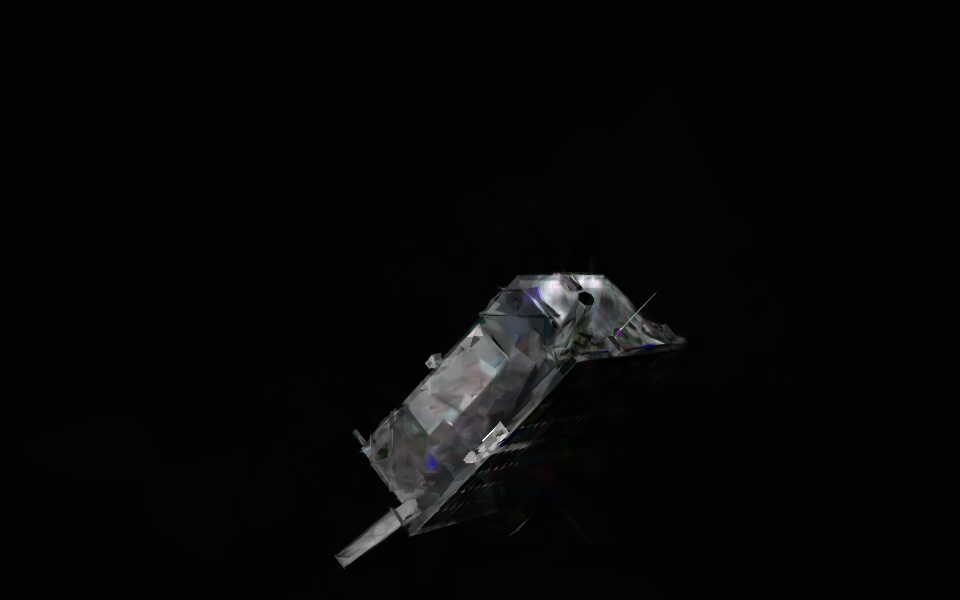} \\
      & \includegraphics[width=0.1\textwidth]{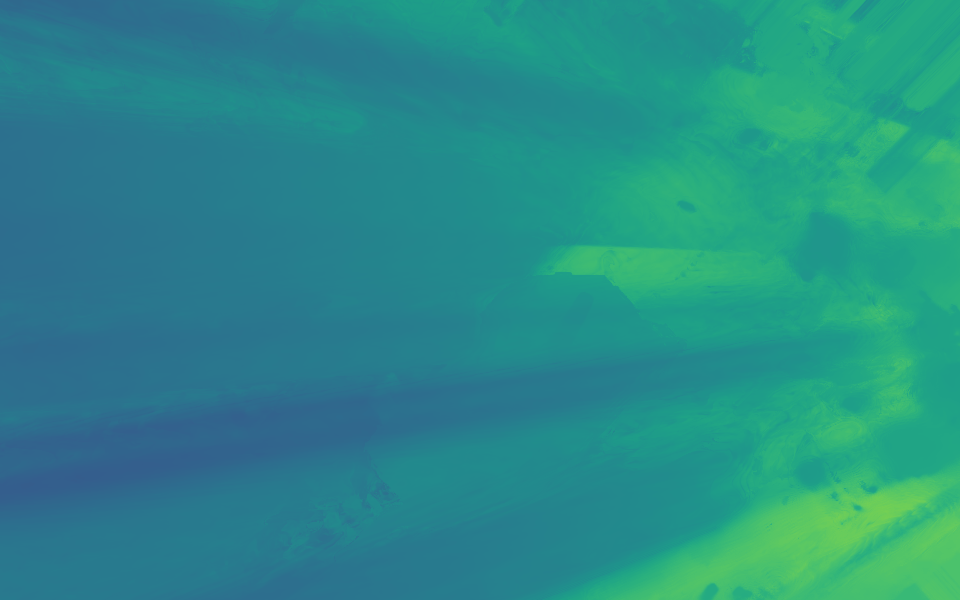} & \includegraphics[width=0.1\textwidth]{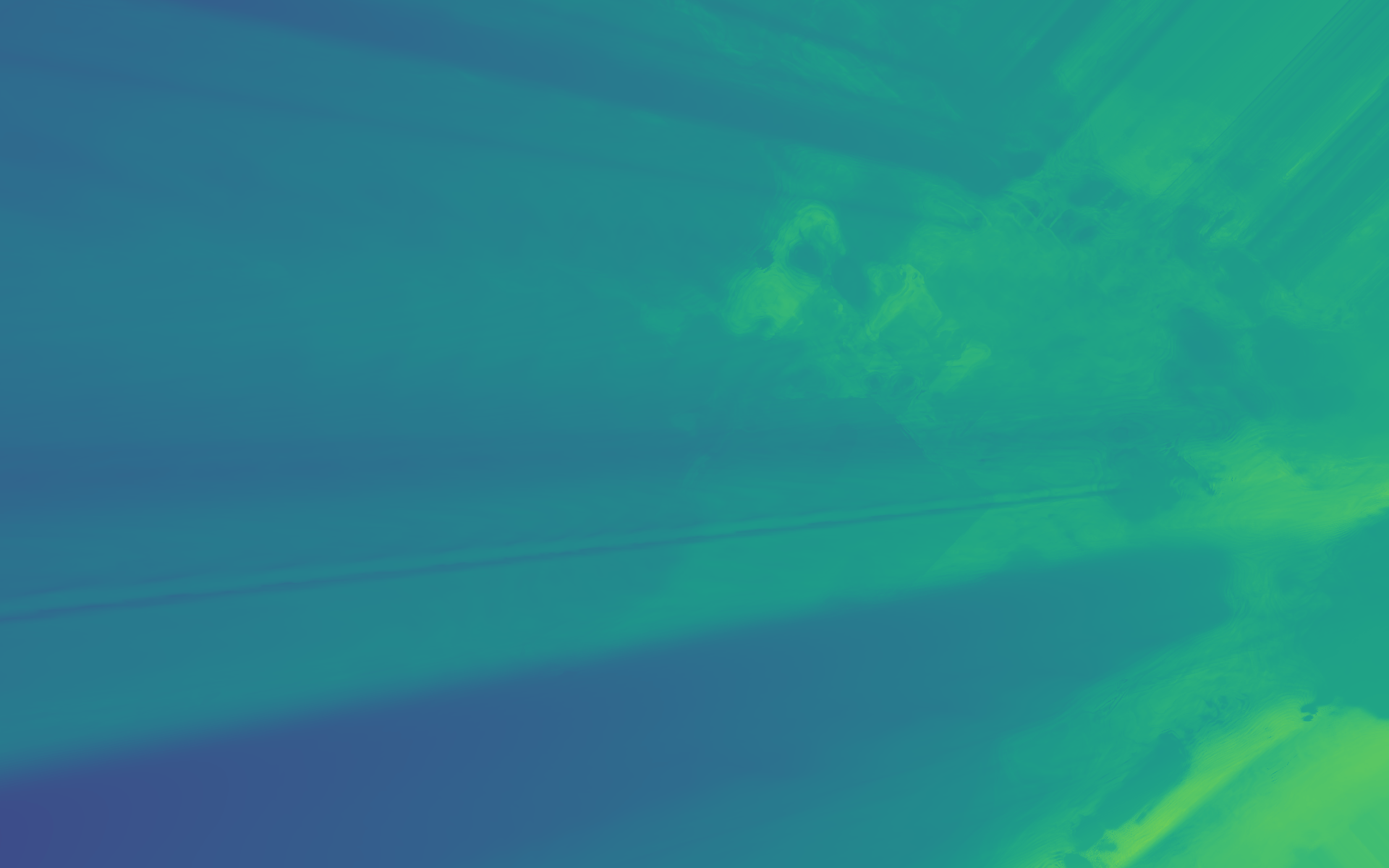} & \includegraphics[width=0.1\textwidth]{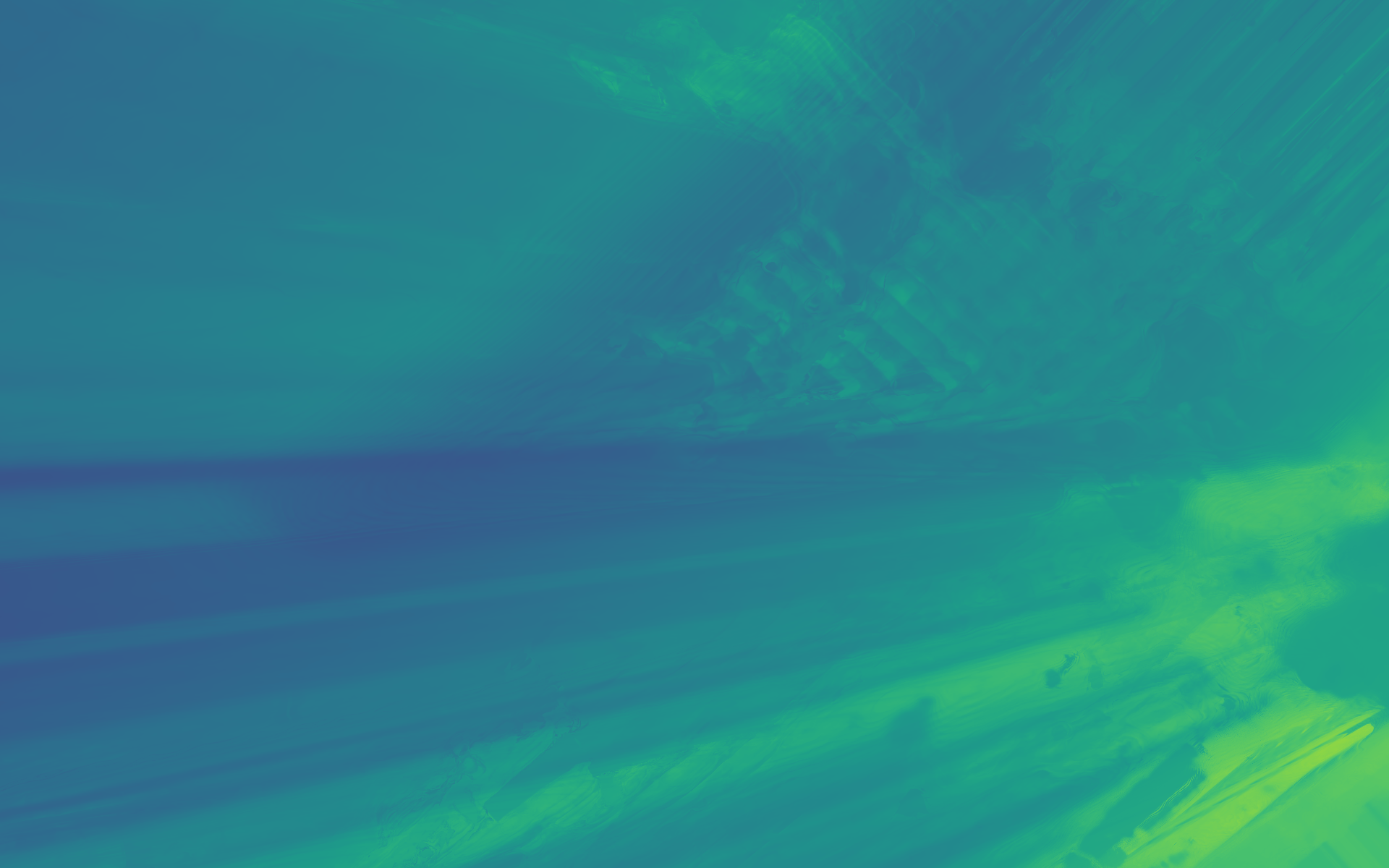} & \includegraphics[width=0.1\textwidth]{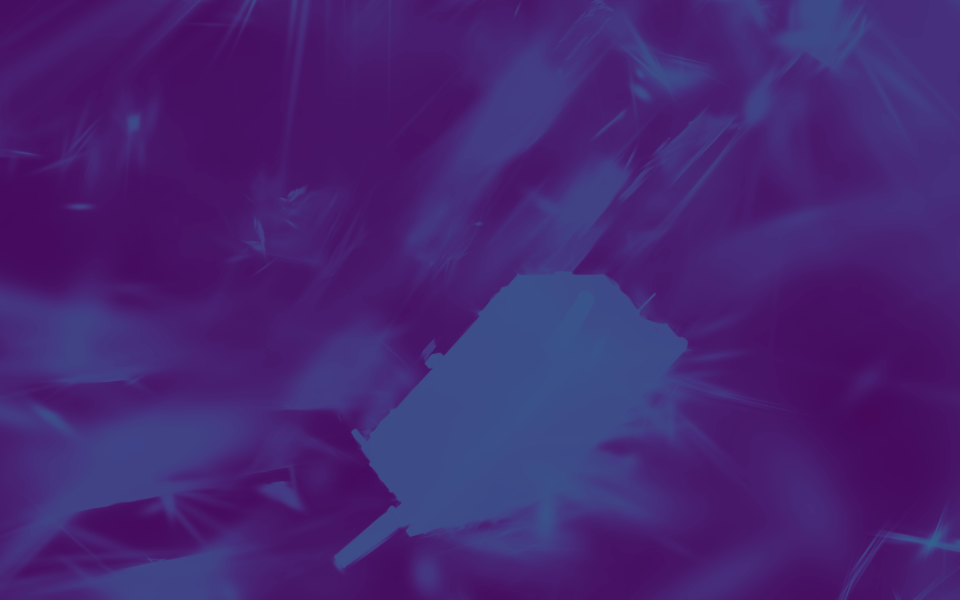} & \includegraphics[width=0.1\textwidth]{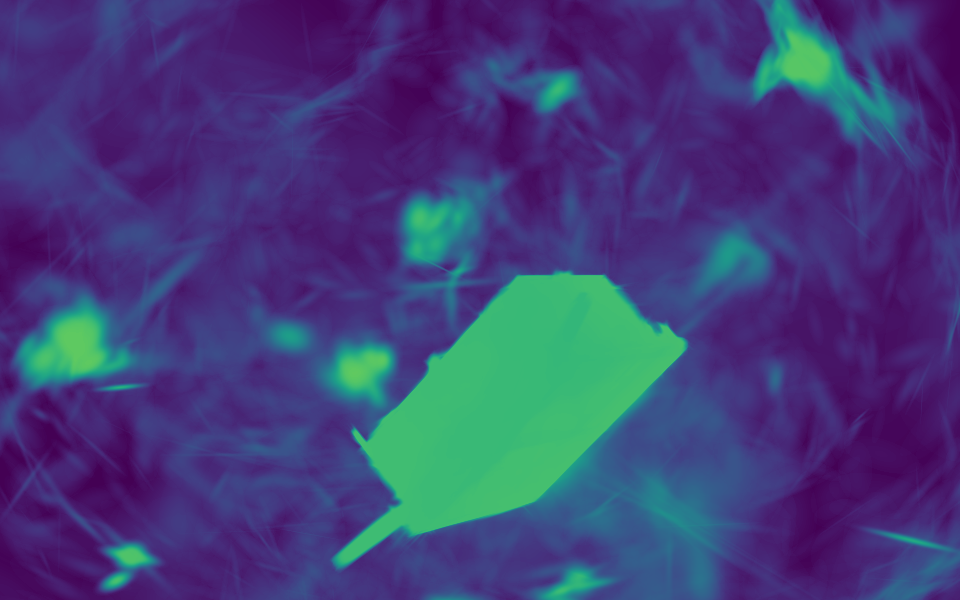} & \includegraphics[width=0.1\textwidth]{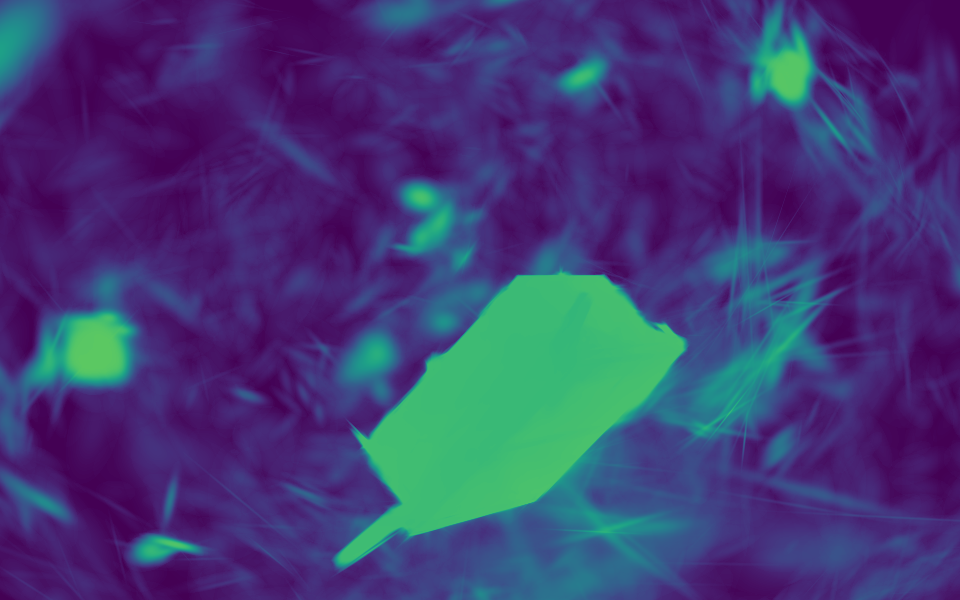} & \includegraphics[width=0.1\textwidth]{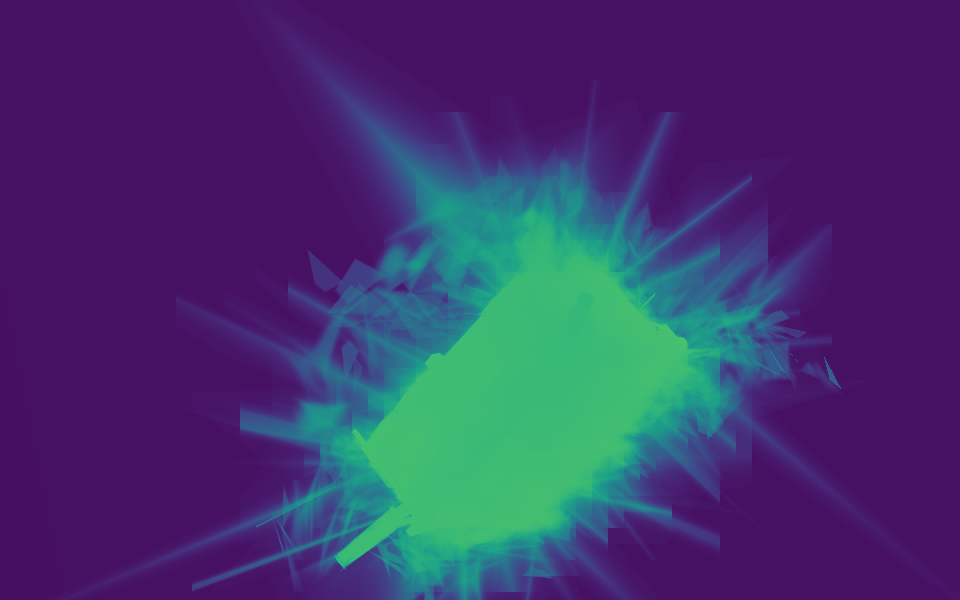} & \includegraphics[width=0.1\textwidth]{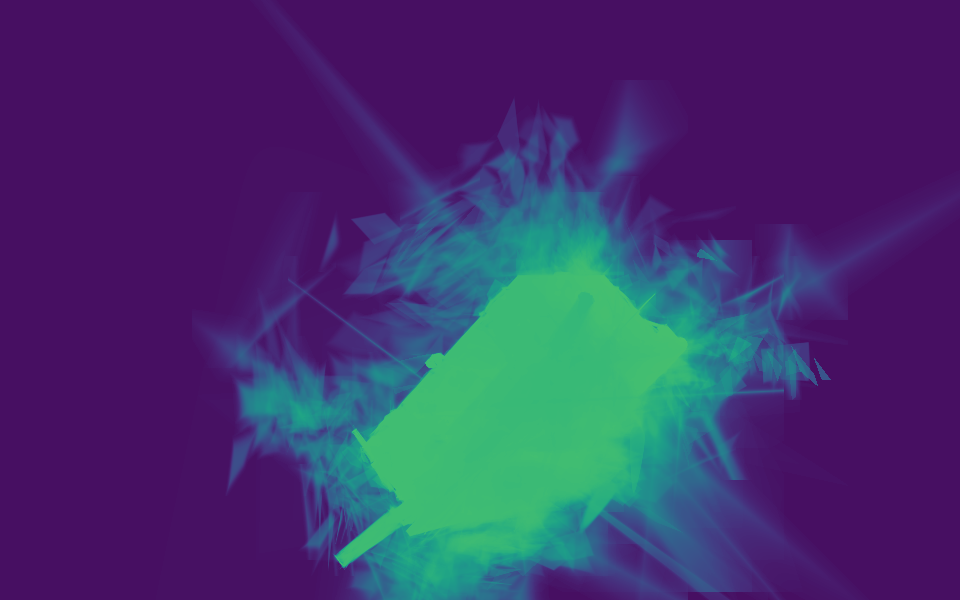} & \includegraphics[width=0.1\textwidth]{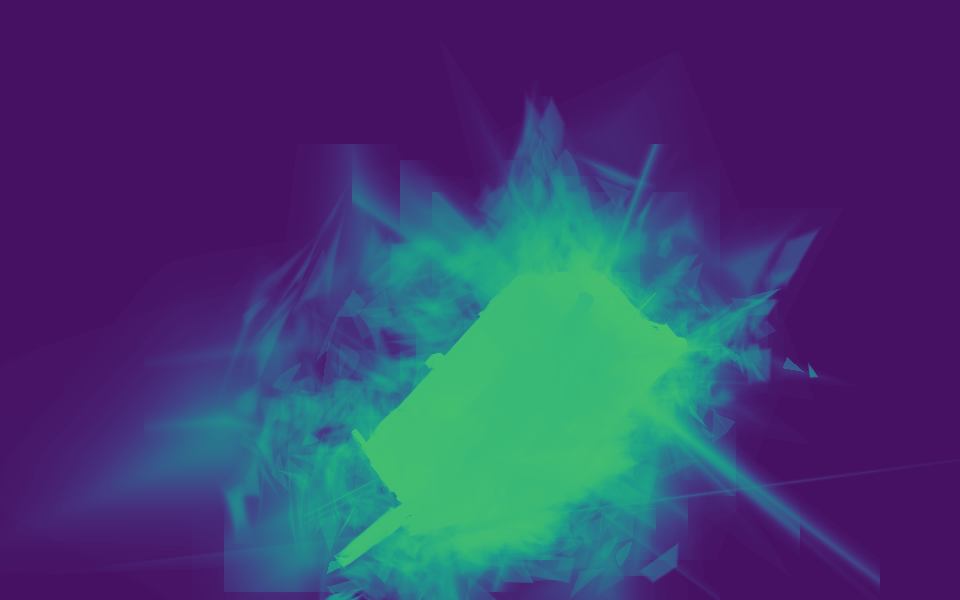} \\
      \bottomrule
   \end{tabular}
   \label{tab:qualitative}
\end{table*}

\subsection{Qualitative Performance}
The qualitative results in Table~\ref{tab:qualitative} highlight differences that are less evident in the quantitative metrics. Unsurprisingly, K-Planes performs the worst: the depth images are filled with spurious geometry, making the spacecraft barely visible, while the renderings show clear artifacts.

For GS, depth images reveal floaters surrounding the satellite across all variants. While appearance embeddings have little geometric effect, they reduce the number of primitives required, which in turn decreases the amount of floaters compared to \emph{vanilla} GS. The two \emph{wild} GS variants differ only at rendering: embeddings optimized with ground truth mask floaters more effectively than random embeddings, though at the expense of leakage.

The imperfections in CS take a different form. Instead of floaters, convex primitives extend slightly beyond the spacecraft, leading to over-coverage rather than isolated clutter. Notably, no floaters appear in the empty space around the object, which could make CS more suitable for space applications where false-positive detections are critical. Photometrically, the CS variants are nearly indistinguishable, consistent with the quantitative results.

Across all methods, the geometry of the spacecraft itself remains largely stable; what changes is the treatment of empty space. Appearance embeddings primarily act as a rendering aid, masking some floaters or color mismatches, but they do not improve the underlying object geometric fidelity.  

\begin{table*}
   \centering
   \caption{Convergence and training time analysis when training the models on a single NVIDIA TeslaA10 GPU. \colorbox{firstcolor}{First}, \colorbox{secondcolor}{second} and \colorbox{thirdcolor}{third} best results are highlighted. * corresponds to appearance embeddings trained on the ground truth image, without corresponds to random appearance embeddings.}
   \begin{tabular}{c||c|c|c||c|c|c||c}
      \toprule
      & \multicolumn{3}{c||}{\textbf{30k training its}} & \multicolumn{3}{c||}{\textbf{60k training its}} & \\
      \midrule
      \textbf{Model} & \textbf{IoU $\uparrow$} & \textbf{PSNR $\uparrow$} & \textbf{\# Parameters $\downarrow$} & \textbf{IoU $\uparrow$} & \textbf{PSNR $\uparrow$} & \textbf{\# Parameters $\downarrow$} & \textbf{Time (minutes) $\downarrow$} \\
      \midrule
      \textbf{VKP} & 0.19 & 19.94 & 33 935 378 & 0.18 & 19.85 & 33 935 378 & 103 \\
      \textbf{WKP} & 0.17 & 19.70 & 33 940 626 & 0.17 & 18.63 & 33 940 626 & 124 \\
      \textbf{WKP*} & 0.17 & 21.05 & 33 940 626 & 0.17 & 20.70 & 33 940 626 & 189 \\
      \textbf{VGS} & \cellcolor{firstcolor}0.45 & \cellcolor{thirdcolor}22.48 & 5 137 444 & \cellcolor{thirdcolor}0.48 & 22.23 & 4 298 150 & \cellcolor{firstcolor}18 \\
      \textbf{WGS} & 0.19 & 19.04 & 1 869 002 & \cellcolor{secondcolor}0.53 & 20.86 & 2 113 574 & \cellcolor{secondcolor}32 \\
      \textbf{WGS*} & 0.20 & 20.82 & 1 769 678 & \cellcolor{firstcolor}0.56 & \cellcolor{firstcolor}25.22 & 2 089 989 & \cellcolor{secondcolor}32 \\
      \textbf{VCS} & \cellcolor{thirdcolor}0.39 & \cellcolor{secondcolor}22.81 & \cellcolor{thirdcolor}1 050 387 & 0.37 & \cellcolor{thirdcolor}22.50 & \cellcolor{thirdcolor}461 127 & \cellcolor{thirdcolor}40 \\
      \textbf{WCS} & \cellcolor{secondcolor}0.40 & 22.37 & \cellcolor{firstcolor}1 002 570 & 0.40 & 21.75 & \cellcolor{firstcolor}418 002 & 44 \\
      \textbf{WCS*} & 0.33 & \cellcolor{firstcolor}23.43 & \cellcolor{secondcolor}1 034 172 & 0.32 & \cellcolor{secondcolor}23.06 & \cellcolor{secondcolor}437 805 & 44 \\
      \bottomrule
   \end{tabular}
   \label{tab:convergence}
\end{table*}

\subsection{Training Time and Convergence}
A final metric of interest in space applications is training time. Although there are typically no strict real-time constraints—since spacecraft maneuvers can span long durations \cite{space_long_operations}, efficiency remains desirable. Table~\ref{tab:convergence} reports training times on a Tesla A10 GPU using the authors' original codebases \cite{kplanes,gaussiansplatting,wildgaussians,convexsplatting} with minor adjustments for the SPEED+ dataset. As expected, K-Planes is significantly slower than the explicit methods due to the costly rendering process: each pixel requires multiple MLP evaluations, while explicit methods rely on a much faster affine transformation.  

To complement these fixed-duration timings, we also inspect convergence behavior by comparing IoU, PSNR, and parameter counts at 30k and 60k iterations. Interestingly, K-Planes already saturates at 30k iterations, but at a poor overall quality — suggesting that more training does not help implicit methods close the gap. By contrast, GS continues to improve in both IoU and PSNR during the second half of training, while the number of parameters remains roughly stable (with a slight increase).  

For CS, performance in IoU and PSNR stabilizes earlier, but the representation becomes much more compact: the number of parameters (and thus primitives) roughly halves between 30k and 60k iterations. This highlights a unique strength of convex splatting: even if raw accuracy plateaus, continued training yields a more efficient, lightweight representation, which is especially valuable in compute- and memory-limited onboard environments.

\paragraph{Summary of Results}
Across all evaluations, a consistent pattern emerges. Implicit methods such as K-Planes converge quickly but at poor geometric and photometric quality, while also incurring heavy computational costs. Explicit methods clearly outperform them: GS achieves the best overall accuracy, though at the price of more primitives, whereas CS yields a more compact representation with slightly reduced fidelity. Appearance embeddings help photometric quality and reduce rendering artifacts but do not improve the underlying geometry. Finally, training-time analysis shows that GS continues to benefit from longer training, while CS converges in accuracy earlier but compresses into a more efficient representation over time. Together, these results highlight an explicit trade-off between accuracy and efficiency, which has direct implications for downstream space robotics applications.

\subsection{Ablation}
\label{sec:ablation}
\begin{table}
   \centering
   \caption{Geometric and photometric metrics when training CS using the hybrid scheme.}
   \begin{tabular}{c|c||c|c}
      \toprule
      \multicolumn{2}{c||}{\textbf{Photometric}} & \multicolumn{2}{c}{\textbf{Geometric}} \\
      \midrule
      \textbf{PSNR $\uparrow$} & 23.12 & \textbf{IoU $\uparrow$} & 0.60 \\
      \textbf{SSIM $\uparrow$} & 0.85 & \textbf{TPR $\uparrow$} & 1.00 \\
      \textbf{LPIPS $\downarrow$} & 0.49 & \textbf{FPR $\downarrow$} & 0.11 \\
       &  & \textbf{FDR $\downarrow$} & 0.40 \\
       &  & \textbf{\# Parameters $\downarrow$} & 137 034 (1 986) \\
      \bottomrule
   \end{tabular}
   \label{tab:hybrid}
\end{table}

\paragraph{Hybrid training of convexes}
Augmenting \emph{vanilla} primitives with appearance embeddings reduces floaters, but the effect differs between methods: GS requires fewer primitives, whereas CS appears to repurpose some noise primitives for geometry. To combine their advantages — efficient geometry from \emph{vanilla} convexes and reduced clutter from \emph{wild} convexes — we propose a two-stage training scheme: train \emph{vanilla} convexes for 30k iterations, then introduce appearance embeddings for an additional 30k. As shown in Table~\ref{tab:hybrid}, this approach matches the best models photometrically while surpassing them geometrically, achieving both higher fidelity and parameter efficiency. The hybrid strategy thus represents a promising step toward compact and functional 3D reconstructions for space applications, though further work is needed to assess whether the drop in image quality affects downstream interaction tasks.

\begin{table*}
   \centering
   \caption{Performance of the different models on different amount of training images, averaged over 20 test images. \colorbox{firstcolor}{First}, \colorbox{secondcolor}{second} and \colorbox{thirdcolor}{third} best results are highlighted.}
   \begin{tabular}{c||c|c|c||c|c|c}
      \toprule
      & \multicolumn{3}{c||}{\textbf{20 training images}} & \multicolumn{3}{c}{\textbf{500 training images}} \\
      \midrule
      \textbf{Model} & \textbf{IoU $\uparrow$} & \textbf{PSNR $\uparrow$} & \textbf{\# Parameters $\downarrow$} & \textbf{IoU $\uparrow$} & \textbf{PSNR $\uparrow$} & \textbf{\# Parameters $\downarrow$} \\
      \midrule
      \textbf{VKP} & 0.21 & 18.08 & 33 935 378 & 0.19 & 21.71 & 33 935 378 \\
      \textbf{WKP} & 0.19 & 16.88 & 33 938 066 & 0.20 & 21.65 & 33 953 426 \\
      \textbf{VGS} & \cellcolor{secondcolor}0.53 & \cellcolor{thirdcolor}19.09 & 8 294 360 & \cellcolor{firstcolor}0.83 & \cellcolor{firstcolor}25.72 & 4 082 948 \\
      \textbf{WGS} & \cellcolor{firstcolor}0.55 & \cellcolor{secondcolor}19.26 & \cellcolor{thirdcolor}1 647 392 & 0.56 & 23.02 & \cellcolor{thirdcolor}1 587 406 \\
      \textbf{VCS} & \cellcolor{thirdcolor}0.44 & 18.69 & \cellcolor{firstcolor}197 961 & \cellcolor{thirdcolor}0.58 & \cellcolor{secondcolor}25.28 & \cellcolor{secondcolor}294 975 \\
      \textbf{WCS} & 0.41 & \cellcolor{firstcolor}19.99 & \cellcolor{secondcolor}215 073 & \cellcolor{secondcolor}0.78 & \cellcolor{thirdcolor}24.95 & \cellcolor{firstcolor}232 185 \\
      \bottomrule
   \end{tabular}
   \label{tab:data_ablation}
\end{table*}

\paragraph{Data Efficiency}
We further evaluate performance as a function of training set size (20 vs. 500 images), reporting IoU, PSNR, and parameter count in Table~\ref{tab:data_ablation}. K-Planes lags behind the explicit methods even with few images, and additional data fails to improve its geometry, indicating an upper bound on accuracy. Explicit methods, by contrast, improve consistently in both geometric and photometric metrics as more data becomes available, with GS outperforming CS at a higher parameter cost. Notably, appearance embeddings are most beneficial in low-data regimes, where they help explain lighting variations. With abundant data (500 images), however, their effect becomes less predictable, as the dataset itself absorbs much of the variation. This highlights a trade-off: embeddings serve as a substitute for missing data, but their utility in learning geometry diminishes as training sets grow.

Overall, these ablation studies underline that explicit representations — particularly when combined with hybrid training or tuned to the data regime — offer a flexible trade-off between efficiency, fidelity, and robustness, reinforcing their suitability for space robotics applications.

\section{Conclusion}
\label{sec:conclusion}
This work presented the first systematic comparison of implicit and explicit neural scene representations for space robotics. Using the SPEED+ dataset, we showed that implicit methods such as K-Planes struggle both geometrically and computationally, while explicit methods achieve higher fidelity and efficiency. Between the explicit approaches, Gaussian splatting provides the best reconstruction quality at the cost of higher parameter counts, whereas convex splatting offers a more compact and clutter-free representation, making it well suited for safety-critical applications. Appearance embeddings mainly improve photometric quality, with limited impact on geometry.  

Overall, our results identify explicit methods — and convex splatting in particular — as promising candidates for onboard deployment in space, combining accuracy with efficiency.

\subsection{Future Work}
We identify two avenues for improving the geometric fidelity of monocular reconstructions.  

First, the expressiveness of primitives could be increased by introducing textured primitives. Appearance embeddings already account for lighting variation, but primitives still share a uniform color. This forces multiple primitives to represent surfaces with varying radiance. Textured primitives \cite{texturegs, texturedgaussians, gstex} can decouple geometry and appearance more effectively, though at the cost of a larger per-primitive parameter count.  

Second, methods beyond view-synthesis-oriented approaches may be explored, such as signed distance functions \cite{deepsdf} or convolutional occupancy networks \cite{conv_occ_netw}. These models optimize for geometry directly rather than relying solely on photometric loss. However, they require specialized training data such as ground-truth meshes, which is costly or infeasible compared to monocular imagery.

\section*{Acknowledgments}
Elias De Smijter is funded by the Wal4XR Walloon Region project. C. De Vleeschouwer is a research director of the Fonds de la Recherche Scientifique - FNRS.

\bibliographystyle{unsrt}
\bibliography{references}

\end{document}